\PassOptionsToPackage{dvipsnames,table }{xcolor}
\documentclass[sigconf]{acmart}
\usepackage{hyperref} 
\usepackage{url} 
\usepackage{enumitem} 
\usepackage{graphicx} 
\graphicspath{ {./images/} } 
\usepackage{wrapfig}  
\usepackage{subfigure} 
\usepackage{array}  
\usepackage{arydshln} 
\usepackage{tabularx} 
\usepackage{multirow} 
\usepackage{wrapfig}
\usepackage{amsmath}
\usepackage{CJKutf8}
\usepackage{diagbox}
\usepackage{balance}
\AtBeginDocument{%
\providecommand\BibTeX{{%
	\normalfont B\kern-0.5em{\scshape i\kern-0.25em b}\kern-0.8em\TeX}}}

\copyrightyear{2023}
\acmYear{2023}
\setcopyright{acmlicensed}\acmConference[CIKM '23]{Proceedings of the 32nd
ACM International Conference on Information and Knowledge
Management}{October 21--25, 2023}{Birmingham, United Kingdom}
\acmBooktitle{Proceedings of the 32nd ACM International Conference on
Information and Knowledge Management (CIKM '23), October 21--25, 2023,
Birmingham, United Kingdom}
\acmPrice{15.00}
\begin{document}

%%
%% The "title" command has an optional parameter,
%% allowing the author to define a "short title" to be used in page headers.
\title{Deep Generative Imputation Model for Missing Not At Random Data}

%%
%% The "author" command and its associated commands are used to define
%% the authors and their affiliations.
%% Of note is the shared affiliation of the first two authors, and the
%% "authornote" and "authornotemark" commands
%% used to denote shared contribution to the research.
\author{Jialei Chen}
\orcid{0009-0003-9861-7788}
\affiliation{%
\institution{MIC Lab,\\Department of Computer Science and Technology,\\Jilin University}
% \institution{Department of Computer Science and Technology}
\city{Changchun}
\country{China}
}
\email{chenjl21@mails.jlu.edu.cn}

\author{Yuanbo Xu}
\orcid{0000-0001-8370-5011}
\authornote{corresponding author}
\affiliation{%
\institution{MIC Lab,\\Department of Computer Science and Technology,\\Jilin University}
% \institution{Department of Computer Science and Technology}
\city{Changchun}
\country{China}
}
\email{yuanbox@jlu.edu.cn}

\author{Pengyang Wang}
\orcid{0000-0003-3961-5523}
\affiliation{%
\institution{SKL-IOTSC, \\Department of Computer and Information Science, \\University of Macau}
\city{Macao}
\country{China}
}
\email{pywang@um.edu.mo}

\author{Yongjian Yang}
\orcid{0000-0002-0056-3626}
\affiliation{%
\institution{MIC Lab,\\Department of Computer Science and Technology,\\Jilin University}
% \institution{Department of Computer Science and Technology}
\city{Changchun}
\country{China}
}
\email{yyj@jlu.edu.cn}

\renewcommand{\shortauthors}{Chen, et al.}

% \author{Anonymous authors}
% \affiliation{
% \institution{Paper under double-blind review}
% \country{~ }
% }
% % \email{ }
% \renewcommand{\shortauthors}{Anonymous Author, et al.}

%%
%% The abstract is a short summary of the work to be presented in the
%% article.
\begin{abstract}
Data analysis usually suffers from the Missing Not At Random (MNAR) problem, where the cause of the value missing is not fully observed. 
Compared to the naive Missing Completely At Random (MCAR) problem, it is more in line with the realistic scenario whereas more complex and challenging. 
Existing statistical methods model the MNAR mechanism by different decomposition of the joint distribution of the complete data and the missing mask. But we empirically find that directly incorporating these statistical methods into deep generative models is sub-optimal. Specifically, it would neglect the confidence of the reconstructed mask during the MNAR imputation process, which leads to insufficient information extraction and less-guaranteed imputation quality. In this paper, we revisit the MNAR problem from a novel perspective that the complete data and missing mask are two modalities of incomplete data on an equal footing. Along with this line, we put forward a generative-model-specific joint probability decomposition method, \textit{conjunction model}, to represent the distributions of two modalities in parallel and extract sufficient information from both complete data and missing mask. Taking a step further, we exploit a deep generative imputation model, namely GNR, to process the real-world missing mechanism in the latent space and concurrently impute the incomplete data and reconstruct the missing mask. The experimental results show that our GNR surpasses state-of-the-art MNAR baselines with significant margins (averagely improved from 9.9\% to 18.8\% in RMSE) and always gives a better mask reconstruction accuracy which makes the imputation more principle.
% A ubiquitous source of nuisance in data analysis is missing data, in which the Missing Not At Random (MNAR) problem is more realistic. Missing data can be split into two modalities, missing-data value ($\mX$) and missing-data mask ($\mM$), mostly bridged through an MNAR mechanism. The MNAR mechanism is easily overlooked and tricky to be handled, generally resulting in biased imputation values. Explicitly modeling the joint distribution of $\mX$ and $\mM$ is feasible to model the MNAR mechanism, but few methods have been attempted. In this work, we propose a new approach to explicitly model the joint distribution, named \textit{fusion factorization model}. Underpinning our approach is the assumption that $\mX$ and $\mM$ are two equal modalities of missing-data multimodality. While the prevailing idea is that $\mX$ contains all the information of $\mM$ and is regarded as the cause of $\mM$. Moreover, the credibility of reconstructed $\mM$ is generally neglected in the imputation process, which makes imputation less guaranteed. We propose a general, scalable deep generative model, IMP-VAE, which has the ability to master the missing mechanism by restoring $\mX$ and reconstructing $\mM$ simultaneously. Our method demonstrates a clear advantage to mainstream baselines in various kinds of datasets and missing scenarios.
% Our setup trades off the flexibility of the missing model and the distortion it induces into the data model when the missing mechanism is MCAR.
\end{abstract}

%%
%% The code below is generated by the tool at http://dl.acm.org/ccs.cfm.
%% Please copy and paste the code instead of the example below.
%%


%%
%% Keywords. The author(s) should pick words that accurately describe
%% the work being presented. Separate the keywords with commas.

%% A "teaser" image appears between the author and the affiliation
%% information and the body of the document, and typically spans the
%% page.
% \begin{teaserfigure}
%   \includegraphics[width=\textwidth]{sampleteaser}
%   \caption{Seattle Mariners at Spring Training, 2010.}
%   \Description{Enjoying the baseball game from the third-base
	%   seats. Ichiro Suzuki preparing to bat.}
%   \label{fig:teaser}
% \end{teaserfigure}

% \received{20 February 2007}
% \received[revised]{12 March 2009}
% \received[accepted]{5 June 2009}

%%
%% This command processes the author and affiliation and title
%% information and builds the first part of the formatted document.

\maketitle

% \begin{figure}[t]       
% \centering       \includegraphics[width=0.46\textwidth,trim=342 208 100 197,clip]{images/matrix.pdf}  		 
% \caption{The imputation process of \textit{selection model} and \textit{conjunction model}. $\mathbf{X}$ and $\mathbf{M}$ are the ground truth data. $\mathbf{X}^{\text{obs}}$ is the observed part of $\mathbf{X}$. $\bar{\mathbf{M}}$ is the \textit{fake mask} and $ \hat{\mathbf{M}}$ is the \textit{probability mask}. Given a probability threshold (e.g., 0.5), $\hat{\mathbf{M}}$ can be directly converted to $\mathbf{M}$.} 
% \label{matrix}   
% \end{figure}
% \vspace{-2mm}

\section{Introduction}
\label{intro}
\begin{figure}[t]
\centering
% \vspace{-4mm}
\subfigure[\textit{Yahoo!R3}]{
	\begin{minipage}[b]{0.45\linewidth}
		\includegraphics[width=1\textwidth]{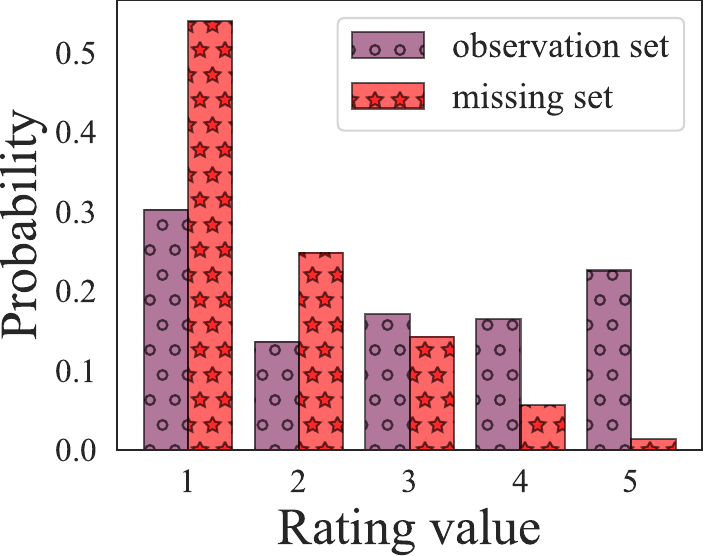}
	\end{minipage}
	
}
\subfigure[\textit{Coat}]{
	\begin{minipage}[b]{0.45\linewidth}
		\includegraphics[width=1\textwidth]{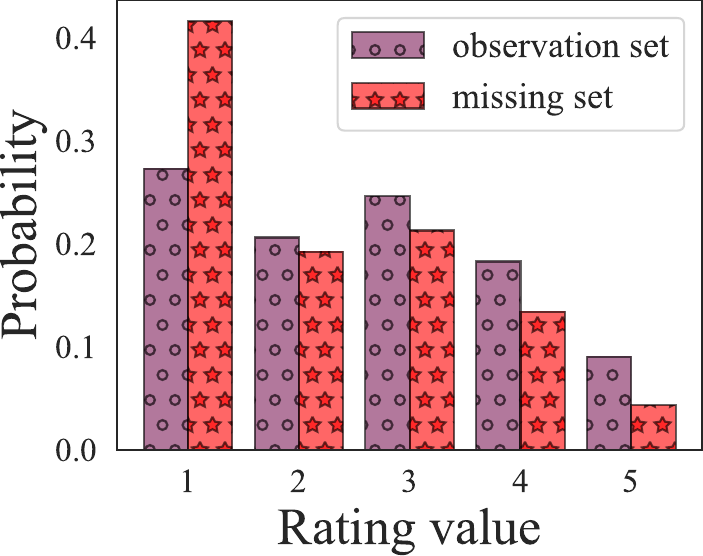}
	\end{minipage}
}
\vspace{-2mm}
\caption{Histogram of the distribution of observed and unobserved (or missing) data shows obvious differences on two five-star rating datasets (\textit{Yahoo!R3} and \textit{Coat}$\text{)}^{\P}$.}
\vspace{-4mm}
\label{data distribution}
\end{figure}

\quad In most real-world scenarios, missing data is an inevitable byproduct during the data-generating process, which severely limits the performance of machine learning methods. Being able to impute the missing data unbiasedly requires correct assumptions about the underlying data-generating process, as well as the missing mechanism in deciding which data is missing.

In general, there are three assumptions of missing mechanisms \citep{rubin1976inference, little2019statistical}. The first assumption is Missing Completely At Random (MCAR), where the probability of data values missing is independent of both the observed and unobserved (or missing) data. In this case, no statistical bias is introduced to the distribution of complete data. The second assumption is Missing At Random (MAR), where the missing mechanism is independent of the value of unobserved data. Under the MAR assumption, maximum likelihood learning can avoid explicit modeling of missing mechanisms by marginalizing missing data. The third assumption is Missing Not At Random (MNAR), where the missing mechanism depends not only on the observed data but also on the missing data.

Due to the strict restrictions of MAR and MCAR assumptions, the MNAR assumption is more realistic and complicated in many real-world scenarios. For example, people are reluctant to rate uninterested items which leaves a large number of low rating values missing from collected data (Rating 1 in Figure~\ref{data distribution}) \citep{chen2023bias}; participants in financial distress are more likely to refuse to complete the survey about financial income. So the cause of missing (hobby or income) can be unobserved and falls under the MNAR mechanism. 
The distribution of observed and missing data usually manifests discrepancies in MNAR data. Ignoring this discrepancy would result in a serious imputation bias and limits the performance of machine learning methods. Thus the MNAR mechanism must be considered.

\begin{figure*}[t]       
\centering       
\vspace{-1mm}
\includegraphics[width=0.95\textwidth,trim=0 20 0 10,clip]{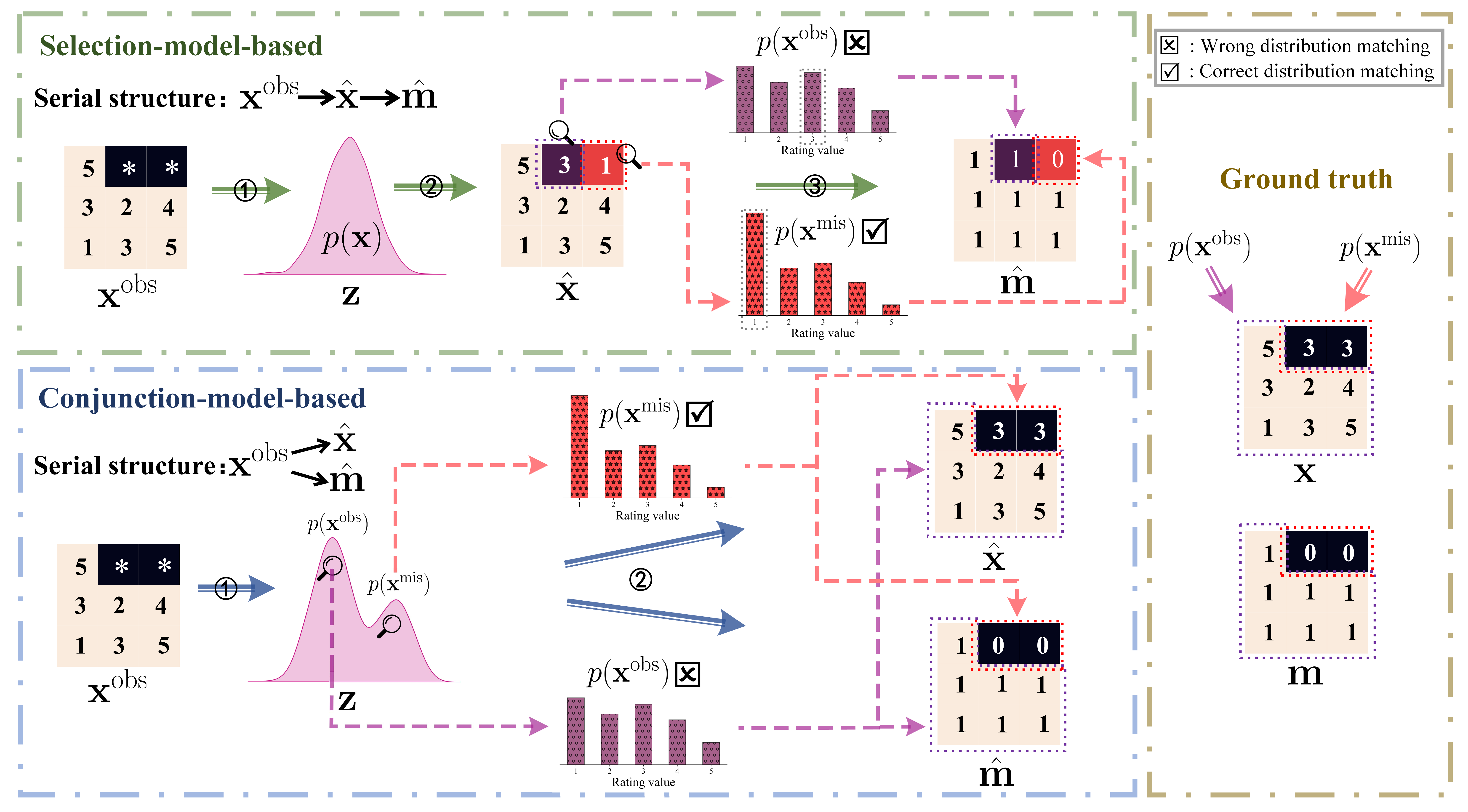} 	 
\vspace{-1mm}
\caption{The procedure of inference with deep generative imputation models based on selection model or conjunction model. The selection-model-based methods use a serial structure to fit the definition of Equation~\ref{4}. The dashed lines are some specific examples of value inference. The reconstructed mask mapping from the imputed data is biased because of the wrong distribution matching. While the conjunction-model-based GNR avoids serial mapping and adopts a parallel structure (Equation~\ref{8}) to extract the information both in the data space and mask space without interfering with each other.} 
\label{model_general}  
\vspace{-2mm}
\end{figure*}

Existing methods for modeling the MNAR mechanism are largely evolved from statistical data analysis by modeling the joint distribution of the complete data and the missing mask. The missing mask is an indicator that reflects whether a value has been observed or unobserved (or missing). According to the different serial decomposition of the joint distribution, they can be divided into the \textit{selection model}, the \textit{pattern-mixture model}, and the \textit{pattern-set mixture model} \citep{little2019statistical}. In recent progress, selection-model-based methods have risen to prominence and derived several deep generative imputation models for MNAR data \citep{ipsen2020not, ma2021identifiable}. The motivation of the selection-model-based methods is to deduce the missing mechanism in the sample space by mapping the data to the mask, with the assumption that the complete data is the cause of values missing. Thus they utilize a serial structure (top of Figure~\ref{model_general}), which involves building an imputation model for presenting the complete data distribution and imputing missing data, followed by reconstructing the mask through a simple mapping of the imputed data.

\renewcommand{\thefootnote}{\fnsymbol{footnote}}
\footnotetext[5]{The two datasets both contain an MNAR set and a small MCAR set. The observation set directly inherits from the MNAR set. Since the MCAR mechanism does not affect the distribution of complete data, we have a unique opportunity to present the distribution of complete data with the MCAR set. We obtain the distribution of the missing set by calculating the difference between the MCAR and MNAR sets.}

However, there are main two challenges with selection-model-based $\text{deep generative imputation models (with serial structure)}^\parallel$:\\
\renewcommand{\thefootnote}{\fnsymbol{footnote}}
\footnotetext[6]{The two additional MNAR modeling methods, which also utilize similar serial structures for mapping data space and mask space, encounter comparable problems.}
$\bullet$ \textbf{Low-quality mask reconstruction.} Some values in observed data and missing data (Rating 2,3,4 in Figure \ref{data distribution}) usually have a similar probability distribution and are hard to distinguish in the complete data. Mapping the complete (or imputed) data to mask has an information bottleneck and can result in wrong distribution matching (top of Figure~\ref{model_general}), which eventually makes the reconstructed mask error as shown in the second row of Figure~\ref{mask}. Since serial structure maps data space to fill a 0 or 1 at some location in the mask, we can abstract this procedure as a classifier. It forces a segmentation of the highly overlapping space of observed and missing data to fit the classification task (mask reconstruction), which is detrimental to the parameter estimation in the data space and eventually leads to biased imputation performance.\\
% $\bullet$ \textbf{Underlying classification problem}. Mapping data space to mask space is like a classifier. It forces the data space to fragment to fit the classification task (mask reconstruction), which is detrimental to the parameter estimation of modeling the data space itself and eventually leads to biased imputation performance.
$\bullet$ \textbf{Neglect of unique information in the missing mask.} Intuitively, the missing mask provides unique information to distinguish observed and missing data (e.g., 0,1 respectively denotes observed and missing). The distribution of the mask can be modeled independently through some carefully-designed structure without misleading the parameter estimate in the data space. Unfortunately, existing methods ignore the unique information and assume that the complete data contains all the information of the mask further utilizing the serial-structure selection model \citep{ipsen2020not, ma2021identifiable, ghalebikesabi2021deep}.

To tackle these challenges, we revisit the relationship between the complete data and the missing mask under the assumption that \textit{they are two modalities of missing-data multimodality}. According to the aforementioned drawbacks of directly mapping complete (or imputed) data to the missing mask, we propose a probability decomposition framework, namely \textit{\textbf{conjunction model}}. It can be well-adopted in the generative models to simultaneously model the distribution of the complete data and the missing mask with a parallel structure, which fully mines the information in the two modalities without interfering with each other.

% To tackle these challenges, we revisit the relationship between the complete data and the mask. Since the complete/imputed data cannot transform into the mask and does not contain all the information about the mask. To extract the rich information in both data space and mask space, we treat complete data and mask as two modalities of missing-data multimodality. Along with this line, we propose the \textit{conjunction model}, a probability decomposition framework designed specifically for deep generative models that explicitly models the joint distribution of complete data and masks for the MNAR problem with parallel structure. This provides a foundation for practical solutions.

Taking a step further, we incorporate the conjunction model into VAEs and put forward a deep \underline{g}enerative imputation model for missing \underline{n}ot at \underline{r}andom data, namely \textbf{GNR}. Specifically, GNR employs a plain encoder to input only the observed data (as obtaining the mask from the observations is intuitive), and two paralleled decoders to reconstruct data space and mask space respectively. The desirable properties of the parallel structure are two-fold: \textit{(i) avoiding the information bottleneck of imputed data during the mask reconstruction; (ii) fully extracting the information about the missing mechanism from not only the data space but also the mask space.} In this way, GNR reconstructs the mask with high credibility and quality which contributes to the rational data imputation. Moreover, we theoretically prove the unbiased evidence lower bound of GNR which ensures promising and stable performance on various tasks.

To summarize, our major contributions can be listed as follows:
\begin{itemize}
    \item We empirically point out that the mask reconstruction in serial-structure methods is sub-optimal, and propose the conjunction model to avoid the information bottleneck and treats the complete data and missing mask as two modalities.
    \item We put forward a parallel-structure deep generative imputation model based on the conjunction model, namely GNR, to simultaneously impute the missing data unbiasedly and reconstruct the missing mask convincingly.
    \item We evaluate the imputation performance of our algorithm on 9 datasets (1 synthetic and 8 real-world). GNR achieves state-of-the-art performance on all validated datasets with multiple missing settings (average surpasses MNAR counterparts by 11.65\%) and consistently reveals accurate mask reconstruction which makes imputation more guaranteed.
    % Towards the mask reconstruction, GNR consistently reveals remarkable and superior accuracy. 
    
\end{itemize}

\section{Problem setting}
\label{background}

\quad Similar to the notations introduced by \citet{ghalebikesabi2021deep} and \citet{yoon2018gain}, Let $ \mathbf{x}= \left(x_1,\ldots,x_d\right)$ be a random variable taking values in the $d$-dimensional full-observed feature space $\mathcal{X} = \mathcal{X}_1 \times \cdots \times \mathcal{X}_d$. To define precisely the observed quantity, we introduce the missing mask vector $\mathbf{m}$: 
\begin{equation}
\begin{split}
	\label{1} 
	m_j &= \left\{ 
	\begin{aligned} &1 && \text{\qquad if}~x_j~\text{is observed,} \\ &0 && \text{\qquad if}~x_j~\text{is missing/unobserved.}\end{aligned} \right.
\end{split}
\end{equation}
Additionally, we introduce the incomplete random variable $\mathbf{x}^{\text{obs}}=\left( x^{\text{obs}}_1,\ldots,x^{\text{obs}}_d \right)\in \mathcal{X} $ which takes values in $ \mathcal{X}=\left( \mathcal{X}_1\cup\{\ast\} \right)\times\cdots\times\left(\mathcal{X}_d\cup\{\ast\}\}\right) $  where $\ast$ is a point not in $\mathcal{X}$ and represents unobserved data points. $ \mathbf{x}^{\text{obs}}$ denotes the set of observed elements of $\mathbf{x}$. $ \mathbf{x}^{\text{obs}}$ can be introduced by
\begin{equation}
% \label{2} 
\mathbf{x}^{\text{obs}} = \ast\odot\left(1-\mathbf{m}\right) + \mathbf{x}\odot\mathbf{m},
\end{equation}
where $\odot$ denotes the Hadamard product. Similarly, we also introduce another random variable $\mathbf{x}^\text{mis}$, which refers to the set of missing elements of $\mathbf{x}$, as follows: 
\begin{equation}
% \label{2} 
\mathbf{x}^{\text{mis}} = \ast\odot\mathbf{m} + \mathbf{x}\odot\left(1-\mathbf{m}\right).
\end{equation}
We can now retrieve $\mathbf{x}$ as: 
\begin{equation}
% \label{2} 
\mathbf{x} = \mathbf{x}^{\text{obs}}\odot\mathbf{m}+\mathbf{x}^\text{mis}\odot\left(1-\mathbf{m}\right).
\end{equation}

% We assume that we are given $n$ i.i.d. samples of $\mathbf{x}^\text{obs}$. The dataset collected by the deployed system can be defined as $ \mathcal{D}=\left\{\mathbf{x}^i\right\}_{i=1}^n$, or $ \mathcal{D} = \left\{\mathbf{x}^{\text{obs}, i},\mathbf{m}^i\right\}_{i=1}^n$. 

Suppose that we have the underlying data-generating process, denoted by $p_{\mathcal{D}}\left(\mathbf{x}^{\text{obs}},\mathbf{m}\right)$, from which we can obtain (partially observed) samples $(x^\text{obs},m)\sim p_{ \mathcal{D} }\left(\mathbf{x}^\text{obs},\mathbf{m}\right)$. We assume that we are given $n$ i.i.d. samples of $\mathbf{x}^\text{obs}$. The dataset collected by the deployed system can be defined as $ \mathcal{D} = \left\{\mathbf{x}^{\text{obs}, i}\right\}_{i=1}^n$ or $ \mathcal{D} = \left\{\mathbf{x}^{\text{obs}, i}, \mathbf{m}^{i}\right\}_{i=1}^n$.  We also have a model to be optimized, denoted by $p_{\theta,\phi}\left(\mathbf{x},\mathbf{m}\right)$. Our goals can then be described as follows:

$\bullet$ To find a new framework to model the joint distribution of $\mathbf{x}$ and $\mathbf{m}$, which better masters the MNAR mechanism. That is, we wish to learn ideal parameters $\hat{\theta}$ and $\hat{\phi}$ for the parametric model $p_{\theta,\phi}(\mathbf{x},\mathbf{m})$ of a deep generative imputation model depending on a new decomposition of the joint distribution that avoids the previous pitfalls, such that $\displaystyle p_{\hat{\theta},\hat{\phi}}(\mathbf{x},\mathbf{m})=p_{\mathcal{D}}(\mathbf{x},\mathbf{m})$.

$\bullet$ Then, given the appropriate parameters, we are able to perform missing-data imputation by $\displaystyle  \mathbb{E}\! \left[ \mathbf{x}^{\text{mis}}|\mathbf{x}^{\text{obs}},\mathbf{m}\!\right]$. If our parameter estimate is unbiased, then our imputation is also unbiased, i.e.,$ \displaystyle p_{\hat{\theta},\hat{\phi}}\left(\mathbf{x}^{\text{mis}}|\mathbf{x}^{\text{obs}},\mathbf{m}\right)\!=\!p_{\mathcal{D}}\left(\mathbf{x}^{\text{mis}}|\mathbf{x}^{\text{obs}},\mathbf{m}\right)$.

\section{Non-negligible missing mechanism modeling}
\label{Nonignorable}

\quad In general, we are interested in obtaining estimates of the parameters $\theta$ in the data model $p_{\theta}(\mathbf{x})$, via maximum likelihood estimation or Bayesian inference. But the assumptions about the missing mechanism determine the appropriate way to learn the data model. To maximize the likelihood of the parameter based only on observed quantities, the missing data is integrated out from the joint distribution, as:
\begin{equation}  
\label{3}
\arg\mathop{\max}_{\theta}p_\theta\left(\mathbf{x}^{\text{obs}},\mathbf{m}\right)
=\arg\mathop{\max}_{\theta}\int p_\theta\left(\mathbf{x}^{\text{obs}},\mathbf{x}^{\text{mis}},\mathbf{m}\right)d \mathbf{x}^{\text{mis}},
\end{equation}

Reviewing the three types of missing mechanisms, if missing data is MCAR or MAR, we can maximize the maximum likelihood only based on observed data \citep{rubin1976inference}, i.e., $p_\theta\left(\mathbf{x}^{\text{obs}},\mathbf{m}\right) \propto p_\theta\left(\mathbf{x}^{\text{obs}}\right)$. But the above argument does not hold for MNAR data. The missing mechanism cannot be neglected during learning and has to be modeled within a joint distribution framework.

\subsection{Existing MNAR modeling approaches}
\label{sec existing}

\quad \citet{little2019statistical} describe three ways of modeling the joint distribution of $\mathbf{x}$ and $\mathbf{m}$ in MNAR case.

\textit{\textbf{Pattern~mixture~model}} \citep{little1993pattern}, decomposes the joint distribution as Equation~\ref{5}, where $p_\theta\left(\mathbf{m}\right)$ is a Bernoulli distribution and $p_\phi\left(\mathbf{x}|\mathbf{m}\right)$ is a mixture of distributions. The key issue is that it requires specifying the distribution of each missing pattern separately. And the data model is stratified by the different missing scenarios, leading to $2^d$ ($\mathbf{m} \in \mathbb{R}^d$) different conditional data models, which is unacceptable on a high-dimensional categorical variable. So it is not suitable for imputation tasks. Moreover mapping the mask which contains only binary information to the data space, obviously has an information bottleneck, and certainly affects the imputation effect of the model seriously. \citet{collier2020vaes} are interested in learning a good generative model of $\mathbf{x}^{\text{obs}}$ and as a result, propose a special \textit{pattern mixture model} based method.
\begin{equation}     \label{5}   p_{\theta,\phi}\left(\mathbf{x},\mathbf{m}\right) = p_\theta\left(\mathbf{m}\right)p_\phi\left(\mathbf{x}|\mathbf{m}\right).   \end{equation}

\textit{\textbf{Selection~model}}~\citep{heckman1979sample}, is the mainstream probability decomposition of the joint distribution (Equation~\ref{4}). It models the distribution of the data and the incidence of missing data (the mask, $\mathbf{m}$) as a function of $\mathbf{x}$. One of our main baseline methods, not-MIWAE~\citep{ipsen2020not}, is a selection-model-based method that uses VAEs to model $p_\theta\left(\mathbf{x}\right)$ and an additional block (e.g., a single layer neural network) to model $p_\phi\left(\mathbf{m}|\mathbf{x}\right)$ behind VAEs. The empirical results of \citet{ghalebikesabi2021deep} show that Not-MIWAE is less robust to possible missing cases. Moreover, allowing the missing model $p_\phi\left(\mathbf{m}|\mathbf{x}\right)$ to be parameterized by a neural network has the disadvantage that the fit of $p_\theta\left(\mathbf{x}\right)$ suffers from the flexibility of the missing model. In the case that the sparsity of data is high, $p_\phi\left(\mathbf{m}|\mathbf{x}\right)$ dominates the model, and the reconstruction of observed data is despised. 
\begin{equation}    \label{4}  p_{\theta,\phi}\left(\mathbf{x},\mathbf{m}\right) = p_\theta\left(\mathbf{x}\right)p_\phi\left(\mathbf{m}|\mathbf{x}\right). \end{equation}

Furthermore, we find the reconstructed mask is always biased and argue that this pitfall is inevitable. We assume that the model is capable to capture both distributions of observed and missing data. The mask reconstruction is essentially solving a classification problem. However, the difference is that traditional classification tasks are sample-level and can make use of rich information from multiple feature spaces, while mask reconstruction depends on each data point with very little information available. Moreover, the sample points have obvious clustering characteristics by category in high-dimensional feature spaces, while the observed and missing values in the data point space are highly overlapping in distribution with the same domains of values (Rating 1-5) and similar probability to be observed or missing (Rating 2-4) in Figure~\ref{data distribution}. Thus the imputed data cannot be precisely divided into observed data and missing data by a simple mapping.

\textit{\textbf{Pattern-set mixture model}} combines the two methods mentioned above (Equation~\ref{6}) \citep{little1993pattern}. $\mathbf{r}$ is an additional categorical latent variable that clusters the missing patterns into $k$ missing pattern sets. $k$ depends on the missing types where the missing data suffers. Therefore it is difficult to set up an appropriate value for $k$. In a special case where $k=1$, it reduces to the selection model. PSMVAE \citep{ghalebikesabi2021deep} can be categorized as a pattern-set mixture model. The pattern-set mixture model is a clustering variant of the selection model. The pitfalls we analyzed in the \textit{selection model} are also in the pattern-set mixture model. 
\begin{equation}     \label{6}   p_{\lambda,\theta,\phi}\left(\mathbf{x},\mathbf{m},\mathbf{r}\right) = p_\lambda\left(\mathbf{r}\right)p_\theta\left(\mathbf{x}|\mathbf{r}\right)p_\phi\left(\mathbf{m}|\mathbf{x},\mathbf{r}\right).   \end{equation}
\vspace{-3mm}
\subsubsection{\textbf{Serial-structure models}}~\\
\label{Serial model}
\quad The above three statistical methods have similarities in serial-structure modeling. All of them first model a distribution of one modality ($p_\theta\left(\mathbf{x}\right)$, $p_\theta\left(\mathbf{m}\right)$ or $p_\theta\left(\mathbf{x}|\mathbf{r}\right)$), and then utilize the serial structure to model the conditional distribution of the other modality depending on the former ($p_\phi\left(\mathbf{m}|\mathbf{x}\right)$, $p_\phi\left(\mathbf{x}|\mathbf{m}\right)$ or $p_\phi\left(\mathbf{m}|\mathbf{x},\mathbf{r}\right)$). We call the above three methods the \textit{serial models}. Briefly, these serial models cannot extract rich information in both data space and mask space which leads to biased imputation performance. 

\subsection{Parallel-structure conjunction model}
\label{sec model}

\quad Now that $\mathbf{x}$ and $\mathbf{m}$ have their unique information and cannot be transformed between each other. From the perspective of multimodal deep learning, we regard the joint distribution modeling of $\mathbf{x}$ and $\mathbf{m}$ as learning a joint representation of two modalities of missing-data multimodality. Under this assumption, we propose a new generative-model-specific probabilistic decomposition method to model the joint distribution in parameters, as: 
\begin{equation}    
\label{7}  
p_{\theta,\phi}\left(\mathbf{x},\mathbf{m},\mathbf{u}\right) = p_\theta\left(\mathbf{u}\right)p_\phi\left(\mathbf{m},\mathbf{x}|\mathbf{u}\right). 
\end{equation}
The detail is that we assume an auxiliary variable $\mathbf{u}$ for the intermediate fusion layer, which has integrated information of multiple modalities, like a joint representation embedding. The fusion layer allows us to capture complementary information about the missing mechanism, which is not visible in individual modalities themselves. The prior information can be injected through $\mathbf{u}$.
\begin{equation}     
\label{8}   
p_{\theta,\phi}\left(\mathbf{x},\mathbf{m},\mathbf{u}\right) = p_\theta\left(\mathbf{u}\right) p_{\phi_1}\left(\mathbf{x}|\mathbf{u}\right) p_{\phi_2}\left(\mathbf{m}|\mathbf{u}\right). 
\end{equation}

MNAR mechanism has so many cases depending on the different relationships between elements of $\mathbf{x}$ and $\mathbf{m}$. To account for complicated MNAR scenarios and to improve the robustness of model specification, we assume that $\mathbf{x}$ and $\mathbf{m}$ are independent conditional on the joint representation $\mathbf{u}$, which is more reasonable for the two modalities. More specifically, $\theta$ is a shared parameter, but $\phi$ is not. $p_{\phi_1}\left(\mathbf{x}|\mathbf{u}\right)$ and $p_{\phi_2}\left(\mathbf{m}|\mathbf{u}\right)$ are two parameter models with different parameters (i.e., two parallel neural networks in practical). On the other hand, the \textbf{parallel architecture} avoids the nuisance suffered by serial models. The new joint model can be presented as Equation~\ref{8}, namely the \textit{\textbf{conjunction model}}. Depending on the flexibility of parallel structure, the conjunction model can extract the rich information in both data space and mask space and generalize kinds of MNAR scenarios, not just self-masking scenarios.

% \begin{figure}[t]     
% \centering     
% \includegraphics[width=0.6\textwidth,trim=0 0 0 0,clip]{iclr2023/images/模型图.jpg} 
% %trim=左 下 右 上
% \caption{Basic structure of IMP-VAE}     
% \label{fig struction} 
% \end{figure}

\section{GNR: a deep generative imputation model for MNAR data}
\label{sec4}

\begin{figure}[t]     
\centering   
% \vspace{-2mm}
\includegraphics[width=0.6\linewidth,trim=250 150 250 160,clip]{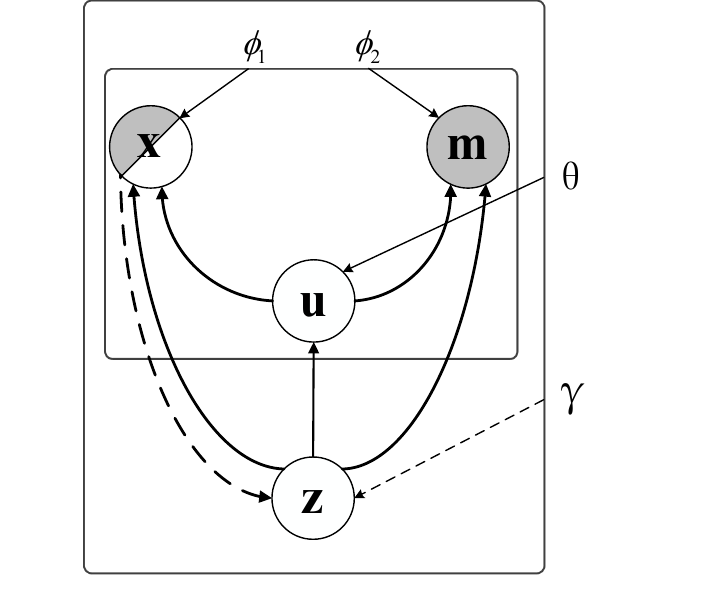}
\vspace{-1mm}
\caption{Graphical representation of GNR.}
\label{fig graphical}
\vspace{-2mm}
\end{figure}

\quad In the previous section, we analyze the existing approaches to model the missing mechanism and propose the conjunction model. However, we still need to derive a practical algorithm that is flexible and compatible with our assumptions. In this section, we propose \textbf{GNR}, a deep generative imputation model for MNAR data based on the conjunction model (Figure~\ref{fig graphical}), which can handle general MNAR scenarios with a large range of missing proportions.

\subsection{Deep generative conjunction model}
\quad We now define a generative model for $p_{\theta,\phi} \left(\mathbf{x}, \mathbf{m}\right)$. The parameters of the joint distribution need to be optimized jointly in an MNAR setting. To maximize the likelihood of the parameters $(\theta,\phi)$ based only on observed quantities, the missing data is integrated out from the joint distribution given by:
\begin{equation}
\label{9}
p_{\theta,\phi}\left(\mathbf{x}^{\text{obs}},\mathbf{m}\right)
=\int p_{\phi}\left(\mathbf{x}^{\text{obs}},\mathbf{x}^{\text{mis}},\mathbf{m}\right) d \mathbf{x}^{\text{mis}}.
\end{equation}
Introducing the conjunction model in the form of Equation~\ref{8}, we can rewrite Equation~\ref{9} as:
\begin{equation}  
\label{10}
\begin{split}
	p_{\theta,\phi}\left(\mathbf{x}^{\text{obs}},\mathbf{m}\right)
	&=
	\int  p_{\phi_1}\left(\mathbf{x}|\mathbf{u}\right)p_{\phi_2}\left(\mathbf{m}|\mathbf{u}\right)p_\theta\left(\mathbf{u}\right) d\mathbf{x}^{\text{mis}}d \mathbf{u},
\end{split}
\end{equation}
where the parameters of the missing-data model and the parameters of the missing-mask model are tied together by $\mathbf{u}$. Then we introduce a latent variable $\mathbf{z}$:
\begin{equation}  
\label{11}
\begin{split}
&p_{\theta,\phi}\left(\mathbf{x}^{\text{obs}},\mathbf{m}\right)\\
	= &\int p_{\phi_1}\left(\mathbf{x}|\mathbf{u},\mathbf{z}\right)p_{\phi_2}\left(\mathbf{m}|\mathbf{u},\mathbf{z}\right)p_\theta\!\left(\mathbf{u}|\mathbf{z}\right)p\left(\mathbf{z}\right)d\mathbf{x}^{\text{mis}} d\mathbf{u} d\mathbf{z}.
\end{split}
\end{equation}

The integral in Equation~\ref{11} is analytically intractable, and direct maximum likelihood methods for learning the parameters $(\theta,\phi)$ are inapplicable. We utilize \textit{importance weighted variational inference} \citep{burda2015importance} to approximate the integral, turning the estimate of a likelihood to the lower bound of the likelihood in an unbiased way. Technically, we introduce an amortized variational inference network $q_\gamma\left(\mathbf{z}|\mathbf{x}^{\text{obs}}\right)$, which comes from a simple family (e.g., the Gaussian family), and its parameter $\gamma$ is learnable through a neural network. The issue is that a neural network cannot deal with variable length input. \citet{nazabal2020handling} use zero imputation (ZI) which first fills the missing value with zero and then feeds the imputed data as input for the inference network. \citet{ma2018eddi} use a permutation invariant set function with the ability to handle the input in variable length. The empirical results show that the effects are similar and we choose the latter. Introducing the variational distribution $q_\gamma\left(\mathbf{z}|\mathbf{x}^{\text{obs}}\right)$, and using the assumption that: $p_{\theta}(\mathbf{x} | \mathbf{u},\mathbf{z})=p_{\theta}\left(\mathbf{x}^{\mathrm{obs}} | \mathbf{u},\mathbf{z}\right)p_{\theta}\left(\mathbf{x}^{\mathrm{mis}} | \mathbf{u},\mathbf{z}\right),$ Equation~\ref{11} is equal to:
\begin{equation}
\label{12}
\begin{split}
	&\log p_{\theta, \phi}\left(\mathbf{x}^{\text {obs}}, \mathbf{m}\right) \\  = &\log \int \frac{p_{\phi_{1}}\left(\mathbf{x}^{\text{obs}} | \mathbf{u}, \mathbf{z}\right) p_{\phi_{2}}\left(\mathbf{m} | \mathbf{u}, \mathbf{z}\right) p_{\theta}\left(\mathbf{u} | \mathbf{z}\right) p\left(\mathbf{z}\right)}{q_{\gamma}\left(\mathbf{z} | \mathbf{x}^{\text {obs }}\right)}\\ &\; q_{\gamma}\left(\mathbf{z} | \mathbf{x}^{\mathrm{obs}}\right)p_{\phi_{1}}\left(\mathbf{x}^{\mathrm{mis}} | \mathbf{u}, \mathbf{z}\right) d \mathbf{x}^{\mathrm{mis}} d \mathbf{u} d \mathbf{z} \\ =& \log \mathbb{E}_{z \sim q_{\gamma}\left(\mathbf{z} | \mathbf{x}^{\mathrm{obs}}\right), x^{\mathrm{mis}} \sim p_{\phi_{1}}\left(\mathbf{x}^{\mathrm{mis}} | \mathbf{u}, \mathbf{z}\right)}\\&~~~~ \left[\frac{p_{\phi_{1}}\left(\mathbf{x}^{\text {obs }} | \mathbf{u}, \mathbf{z}\right) p_{\phi_{2}}\left(\mathbf{m} | \mathbf{u}, \mathbf{z}\right) p_{\theta}\left(\mathbf{u} | \mathbf{z}\right) p\left(\mathbf{z}\right)}{q_{\gamma}\left(\mathbf{z} | \mathbf{x}^{\mathrm{obs}}\right)}\right].
\end{split}
\end{equation}

The prior information can be added to the model by obtaining prior embedding from a prior network and injecting it into the latent space through $\mathbf{u}$. If no extra information can be introduced, we can select the low-sparsity samples or features as the heuristic prior or even omit $\mathbf{u}$ under the assumption that $\mathbf{z}$ has learned all the latent representation we need. Since $\mathbf{u}$ is an auxiliary variable, we do not need to use a large number of samples to estimate its expectation. We replace other expectations inside the logarithm with Monte Carlo estimates. Let 
\begin{equation}
\label{13}
\mathcal{L}_K\left(\theta,\phi,\gamma\right)=\mathbb{E}_{\left\{z^k,x^{\text{mis},k}\right\}_{k\in{\{1,\ldots,K\}}}}
\left[\log \frac{1}{K}\sum\limits_{k=1}^K\omega_{k} \right], 
\end{equation}
where, for all $k \leq K$, $\omega_k$ is the importance weight and calculated as:
\begin{equation}
\label{14}
\omega_k=
\frac{p_{\phi_1}\left(\mathbf{x}^{\text{obs}} | \mathbf{u},\mathbf{z}^k\right) p_{\phi_2}\left(\mathbf{m}| \mathbf{u},\mathbf{z}^k\right) p_{\theta}\left(\mathbf{u}| \mathbf{z}^k\right)p\left(\mathbf{z}^k\right)}
{q_{\gamma}\left(\mathbf{z}^k |\mathbf{x}^{\text{obs}}\right)}.
\end{equation}
$\left\{z^k,x^{\text{mis},k}\right\}$ are $K$ i.i.d. samples from $q_{\gamma}\left(\mathbf{z} |\mathbf{x}^{\text{obs}} \right)$ and $p_{\phi_1}\left(\mathbf{x}^{\text{mis}}|\mathbf{u},\mathbf{z}\right)$ through \textit{reparameterization trick} \citep{kingma2013auto}. Similar to the approach proposed in \citet{ipsen2020not}, we directly give the monotonicity property of $ \mathcal{L}_{K}$ and the convergence to the true likelihood in Equation~\ref{15}. The unbiasedness of the Monte Carlo estimates ensures (via Jensen’s inequality) that the objective is indeed a lower bound of the likelihood.
\begin{equation}
\label{15}
\mathcal{L}_{1}\left(\theta, \phi, \gamma\right) \leq \cdots \leq \mathcal{L}_{K}\left(\theta, \phi, \gamma\right) \underset{K \rightarrow \infty}{\longrightarrow} \log p_{\theta, \phi}\left(\mathbf{x}^{\text{obs}}, \mathbf{m}\right),
\end{equation}
where $\displaystyle \mathcal{L}_{K}\left(\theta, \phi, \gamma\right)$ is an unbiased importance weighted lower bound of $\displaystyle \log p_{\theta, \phi}\left(\mathbf{x}^{\text{obs}}, \mathbf{m}\right)$, named \textbf{\textit{GNR lower bound}}. 

In addition, to handle a wide range of missing scenarios, we introduce a hyper-parameter $\alpha$ for $ \displaystyle p_{\phi_2}\left(\mathbf{m} | \mathbf{u},\mathbf{z}^k\right)$ in Equation~\ref{16}. $\alpha$ handles the trade-off between learning the model $p_{\phi_2}\left(\mathbf{m} | \mathbf{u},\mathbf{z}^k\right)$ that explains the missing mask better and learning the model $p_{\phi_1}\left(\mathbf{x}^{\text{obs}} | \mathbf{u},\mathbf{z}^k\right)$ that explains the observable variables better. When $\alpha=1$,  $ \hat{\omega}$ is equivalent to $\omega$. In a special condition, when $\alpha \to 0$, the model degenerates to a VAE similar to MIWAE~\citep{mattei2019miwae}, which can handle MCAR/MAR data. 
\begin{equation}     
\label{16}     
\hat{\omega}_k=     \frac{p_{\phi_1}\left(\mathbf{x}^{\text{obs}} | \mathbf{u},\mathbf{z}^k\right) \left[p_{\phi_2}\left(\mathbf{m} | \mathbf{u},\mathbf{z}^k\right)\right]^\alpha p_{\theta}\left(\mathbf{u}| \mathbf{z}^k\right)p(\mathbf{z}^k)}     {q_{\gamma}\left(\mathbf{z}^k |\mathbf{x}^{\text{obs}}\right)} .
\end{equation}

Note that we neither use two encoders to specifically encode $\mathbf{x}$ and $\mathbf{m}$, nor concatenate them to feed one encoder. We do not include the missing mask $\mathbf{m}$ as additional input to $q_\gamma$, as we want GNR to learn a convincing missing mask by itself through the MNAR mechanism. We can optimize the parameters $\theta,\phi,\gamma$ by solving the following optimization problem:
\begin{equation}
\label{17}
\hspace{-2mm}
\theta^\ast,\phi^\ast,\gamma^\ast\!=\!
\arg\mathop{\max}_{\theta,\phi,\gamma} 
\mathbb{E}_{\left(x^{\text{obs}},m\right)\sim p_D\left(\mathbf{x}^{\text{obs}},\mathbf{m}\right)}\mathcal{L}_K\left(\theta,\phi,\gamma,\mathbf{x}^{\text{obs}},\mathbf{m} \right).
\end{equation}

\textbf{Imputation.} Given $\theta^\ast,\phi^\ast,\gamma^\ast$, we can impute the missing data by estimating $\displaystyle \mathbb{E}_{x^{\text{mis}}}\left[ h\left(\mathbf{x}^{\text{mis}}\right)|\mathbf{x}^{\text{obs}},\mathbf{m}\right]$ using self-normalized importance sampling \citep{ipsen2020not}. In the case that the $l_2$ norm is a relevant error metric, $h$ is the identity function:
\begin{equation}
\label{18}
\begin{split}
	\hat{\mathbf{x}}^{\text{mis}}
	&=\mathbb{E}_{x^{\text{mis}}}\!\left[ h\left(\mathbf{x}^{\text{mis}}\right)|\mathbf{x}^{\text{obs}},\mathbf{m}\right]
	\\&\approx \sum\limits_{l=1}^L\frac{\hat{\omega_l}}{\sum_{l=1}^L \hat{\omega}_l}\mathbb{E}_{x^{\text{mis}}}\!\left[ \mathbf{x}^{\text{mis}}|\mathbf{x}^{\text{obs}},\mathbf{m}\right],
\end{split}
\end{equation}
where $L>>K$. It is also possible to perform multiple imputations with the same computations. For this, we generate a set of imputations by using sampling importance resampling and weight them using the weights defined in Equation~\ref{18}. 
% We also demonstrate a supervised version of GNR, named SGNR, in Appendix~\ref{SGNR} to show the scalability of our method.

\begin{table*}[t]
 % \captionsetup{justification=centering}
 \vspace{-3mm}
\caption{Imputation RMSE for synthetic data under multiple missing scenarios. XX\% is related to a probability of a value above the mean missing (MNAR). \%Improv.:percentage of improvement on metrics over the best baseline.}
\vspace{-1mm}
% \resizebox{0.72\textwidth}{!}{
	\setlength{\tabcolsep}{0.035\linewidth}{
		\begin{tabular}{lcccccccc}
			\toprule[1.25pt]
			\textbf{Method} & \textbf{20\% }& \textbf{80\%} & \textbf{100\%} & \textbf{$\star$}     & \textbf{80\%+20\%MCAR} \\
			\midrule[0.75pt]
			
			MIWAE & 1.06 +- 0.03 & 1.45 +- 0.01 & 1.71 +- 0.02 & 1.70 +- 0.02 & 1.27 +- 0.01 \\
			not-MIWAE & \underline{0.94 +- 0.04} & 1.26 +- 0.03 & \underline{1.33 +- 0.03} & 1.38 +- 0.06 & \underline{1.16 +- 0.03} \\
			PSMVAE & 0.98 +- 0.02 & \underline{1.17 +- 0.03} & 1.39 +- 0.03 & \underline{1.31 +- 0.07} & 1.25 +- 0.02 \\
			\textbf{GNR} & \textbf{0.87 +- 0.05} & \textbf{0.91 +- 0.03} & \textbf{1.14 +- 0.04} & \textbf{1.15 +- 0.06} & \textbf{1.04 +- 0.01}  \\
			mean  & 1.05 +- 0.04 & 1.45 +- 0.01 & 1.71 +- 0.01 & 1.71 +- 0.01 & 1.26 +- 0.01  \\
			MICE  & 1.05 +- 0.04 & 1.45 +- 0.01 & 1.71 +- 0.01 & 1.71 +- 0.01 & 1.29 +- 0.04  \\
			MissForest & 1.11 +- 0.05 & 1.49 +- 0.01 & 1.74 +- 0.02 & 1.76 +- 0.01 & 1.36 +- 0.03  \\
			\midrule[0.75pt]
			\%Improv. & +7.4\% & +22.2\% & +14.3\% & +12.2\% & +10.3\%
			\\
			\bottomrule[1.25pt]
			\multicolumn{6}{l}{
				\textbf{$\star$} : if first feature value $ x_1 >  \bar{x}_1$ (feature mean) or second feature value $x_2< \bar{x}_2$, $x_{1}/x_{2}$ will be missing with 100\% probability.
			}
			
	\end{tabular}}
	\label{fig gauss}
 % \vspace{-1mm}
\end{table*}

\section{Experiments}
\label{experiment}

\quad In this section, we quantitatively evaluate the imputation performance of our model and several state-of-the-art approaches on synthetic datasets (Section~\ref{synthetic}), real-world datasets with synthetic missing settings (Section~\ref{UCI}), and real-world MNAR datasets with MCAR test sets (Section \ref{RCT}). The best score for each group is highlighted in \textbf{bold} and the next best is \underline{underlined}. 

\textbf{The experimental setting details.} We first introduce the general settings of GNR and other baselines. GNR is based on the \textit{conjunction model} in Section~\ref{sec4}. The data model ($p_{\phi_1}\!\left(\mathbf{x}|\mathbf{u},\mathbf{z}\right)$) for GNR is parameterized with Gaussian likelihood functions. The missing mask model ($p_{\phi_2}\!\left(\mathbf{x}|\mathbf{u},\mathbf{z}\right)$) uses Bernoulli likelihood and sigmoid activation. All the VAE-based approaches have two hidden layers with 128 nodes for both the encoder and decoder. All neural networks use Tanh activation except for output layers. Gaussian distributions are used as the variational distribution in the latent space. We use the Adam optimizer with a learning rate of 0.001 and train for 10k iterations with a batch size of 128. All VAE-based baselines use importance-weighted VAE objectives with k = 20 importance samples, and L = 1000 is used for estimating the imputation performance. All the code of MF-based methods can be found in \citet{wang2020information}. 

\textbf{Missing settings.} We introduce synthetic missing data as follows: the MNAR data are generated by self-masking in \textbf{half of the features}: $\mathbf{x}_i$ is missing, if $\mathbf{x}_i$ is greater than the feature mean, with some certain probability (e.g., 20\%, 80\%, and 100\%). In this case, the sparsity of a dataset with a $k$ probability of missing is about $k/4$. The MCAR characteristic is introduced by random masking in \textbf{all of the features} to be missing with some certain probability. 

\subsection{Synthetic MNAR dataset}
\label{synthetic}

\begin{table*}[t]
\renewcommand\arraystretch{1.1}
\vspace{-1mm}
\centering
\caption{Imputation RMSE on UCI datasets under different missing scenarios.} 
\vspace{-1mm}
\label{fig UCI}
\resizebox{\textwidth}{!}{%
\begin{tabular}{lccccccccccccccclll}
\toprule[1.25pt]     \multicolumn{1}{c}{\multirow{2}{*}{\diagbox{\footnotesize Method}{\footnotesize Dataset}}}                      & \multicolumn{3}{c}{\textit{\textbf{white wine}}}                         & \multicolumn{3}{c}{\textit{\textbf{abalone}}}                           & \multicolumn{3}{c}{\textit{\textbf{banknote}}}                           & \multicolumn{3}{c}{\textit{\textbf{Yeast}}}                           & \multicolumn{3}{c}{\textit{\textbf{red wine}}}                   & \multicolumn{3}{c}{\textit{\textbf{concrete}}} \\
\cmidrule(lr){2-4}  \cmidrule(lr){5-7} \cmidrule(lr){8-10} \cmidrule(lr){11-13}   \cmidrule(lr){14-16}   \cmidrule(lr){17-19} & 20\%             & 80\%             & 100\%            & 20\%             & 80\%             & 100\%            & 20\%             & 80\%             & 100\%            & 20\%             & 80\%          & 100\%            & 20\%          & 80\%          & 100\%         & 20\%          & 80\% & 100\% \\ \midrule[0.75pt]
         MIWAE                                                                                                        & 1.04             & 1.25             & 1.55             & 0.63             & 0.44             & 0.80             & 0.36             & 0.81             & 1.26             & 1.07             & 1.51          & 1.73             & 0.98          & 1.24          & 1.62           & 0.43          & 1.25 & 1.70  \\
not-MIWAE                                                                                                                     & \underline{0.91} & \underline{1.04}             & \underline{1.18} & 0.52 & \underline{0.43} & 0.94 & \underline{0.31}    & 0.46             & 1.06 & \underline{0.97} & 1.28          & 1.52             & \underline{0.90}          & \underline{1.07}          & 1.25          & \underline{0.42}          & 0.81 & 1.32  \\
PSMVAE                                                                                                                        & 0.94             & 1.09 & 1.21             & \underline{0.45}             & 0.47            & \underline{0.74}             & 0.33             & \underline{0.43} & \underline{0.97}             & 0.99             & \underline{1.14}        & \underline{1.28} & 0.92          & 1.19          & \underline{1.21}          & 0.56          & \underline{0.79} & \underline{1.15}  \\
\textbf{GNR}                                                                                                                  & \textbf{0.87}    & \textbf{0.93}    & \textbf{1.02}    & \textbf{0.33}    & \textbf{0.41}    & \textbf{0.47}    & \textbf{0.30} & \textbf{0.40}    & \textbf{0.72}    & \textbf{0.84}    & \textbf{0.87} & \textbf{0.98}    & \textbf{0.84} & \textbf{1.04} & \textbf{1.14} & \textbf{0.40} & \textbf{0.69} & \textbf{1.06} \\ \midrule[0.5pt]      
\%Improv. &+4.4\% &+10.6\% &+13.6\% &+26.7\% &+4.7\% &+36.5\% &+3.2\% &+7.0\% &+25.8\% &+13.4\% &+23.7\% &+23.4\% &+6.7\% &+2.8\% &+5.8\% &+4.8\% &+12.7\% &+7.8\%   
\\ \midrule[0.75pt]
         mean                                                                                                          & 1.32             & 1.59             & 1.74             & 1.05             & 1.39             & 1.69             & 1.12             & 1.46             & 1.73             & 1.13             & 1.52          & 1.73             & 1.41          & 1.68          & 1.84          & 1.15          & 1.60 & 1.85  \\
MICE                                                                                                                          & 1.05             & 1.21             & 1.41   & 0.54             & 0.58             & \textbf{0.61}             & 0.69             & 0.98             & \textbf{1.41}    & \textbf{1.04}    & \textbf{1.42} & 1.76             & 0.94          & 1.16          & 1.68          & 0.51          & \textbf{0.74} & \textbf{1.69}  \\
MissForest                                                                                                                    & \textbf{0.82}    & \textbf{1.15}    & 1.63             & \textbf{0.53}    & \textbf{0.55}    & 1.32    & \textbf{0.35}    & \textbf{0.74}    & \textbf{1.28}    & 1.05             & 1.43          & \textbf{1.71}    & \textbf{0.81}          & \textbf{1.09}          & \textbf{1.64}          & \textbf{0.32}          & 0.78 & 1.76  \\ \bottomrule[1.25pt] 
\end{tabular}%
}
% \vspace{-2mm}
\end{table*}

% Please add the following required packages to your document preamble:
% \usepackage{booktabs}
% \usepackage{multirow}
% \usepackage{graphicx}

\begin{table*}[t]
\renewcommand\arraystretch{1.3}
\centering
\vspace{-1mm}
\caption{Reconstructed mask accuracy on UCI datasets under different missing scenarios. We only measure features that contain missing values. RANDOM means the accuracy lower bound of the mask which is generated completely randomly.}
\vspace{-1mm}
\label{mask_acc}
\resizebox{\textwidth}{!}{%
\begin{tabular}{@{}lccccccccccccccclll@{}}
\toprule[1.25pt]     \multicolumn{1}{c}{\multirow{2}{*}{\diagbox{\footnotesize Method}{\footnotesize Dataset}}} &
  \multicolumn{3}{c}{\textit{\textbf{white wine}}} &
  \multicolumn{3}{c}{\textit{\textbf{abalone}}} &
  \multicolumn{3}{c}{\textit{\textbf{banknote}}} &
  \multicolumn{3}{c}{\textit{\textbf{yeast}}} &
  \multicolumn{3}{c}{\textit{\textbf{red wine}}} &
  \multicolumn{3}{c}{\textit{\textbf{concrete}}} \\ 
\cmidrule(lr){2-4}  \cmidrule(lr){5-7} \cmidrule(lr){8-10} \cmidrule(lr){11-13}   \cmidrule(lr){14-16}   \cmidrule(lr){17-19} &
  20\% &
  80\% &
  100\% &
  20\% &
  80\% &
  100\% &
  20\% &
  80\% &
  100\% &
  20\% &
  80\% &
  100\% &
  20\% &
  80\% &
  100\% &
  20\% &
  80\% &
  100\% \\ \midrule[0.75pt]
RANDOM       & 82.00\% & 52.00\% & 50.00\% & 82.00\% & 52.00\% & 50.00\% & 82.00\% & 52.00\% & 50.00\% & 82.00\% & 52.00\% & 50.00\% & 82.00\% & 52.00\% & 50.00\% & 82.00\% & 52.00\% & 50.00\% \\
not-MIWAE    & 84.23\% & 68.49\% & 78.52\% & 82.26\% & 74.43\% & 83.53\% & 83.12\% & 75.85\% & 72.48\% & 83.36\% & 56.34\% & 53.27\% & 86.19\% & 73.24\% & 83.53\% & 84.61\% & 70.75\% & 76.16\% \\
\textbf{GNR} & \textbf{93.54\%} & \textbf{93.37\%} & \textbf{96.42\%} & \textbf{87.79\%} & \textbf{88.82\%} & \textbf{98.14\%} & \textbf{84.79\%} & \textbf{91.02\%} & \textbf{93.44\%} & \textbf{94.62\%} & \textbf{89.76\%} & \textbf{92.90\%} & \textbf{88.77\%} & \textbf{95.16\%} & \textbf{95.12\%} & \textbf{95.33\%} &\textbf{96.48\%} & \textbf{96.87\% }\\ \bottomrule[1.25pt]
\end{tabular}%
}
\vspace{-1mm}
\end{table*}

\quad First, to evaluate the superiority of our \textit{conjunction model}, we consider a $4~d$ synthetic MNAR dataset ($\mathbf{x} \in \mathbb{R}^4$), which consists of samples from a multivariate Gaussian distribution and multiple missing cases. If the missing mechanism is MNAR, $\mathbf{x}_1, \mathbf{x}_2$ are self-masking, and $\mathbf{x}_3, \mathbf{x}_4$ are fully observed. We not only model MNAR but also created a version of the data set (the last column of Table~\ref{fig gauss}) where all features are MCAR and half of the features are additionally MNAR. We use 1-dimensional latent space and zero imputation for the simple structure of synthetic data. The mean and standard errors are found over 5 runs.
% We train GNR and baseline models with partially observed data. To evaluate the model performance, we compare GNR with the following baselines: 1), MIWAE~\citep{mattei2019miwae}; 2), not-MIWAE~\citep{ipsen2020not} (MIWAE+\textit{selection model}); 3), PSMVAE~\citep{ghalebikesabi2021deep}; and 4), traditional methods: mean imputation, MICE with Bayesian Ridge regression~\citep{van2011mice}, and MissForest~\citep{stekhoven2012missforest}.
We compare GNR with the following baselines: i) MIWAE~\citep{mattei2019miwae}: a VAE-based model specifically designed for MAR data with Zero Imputation; ii) not-MIWAE~\citep{ipsen2020not}: incorporating selection model into MIWAE for MNAR data; iii) PSMVAE~\citep{ghalebikesabi2021deep}: incorporating pattern-set mixture model into MIWAE; iv) traditional methods: mean imputation, MICE with Bayesian Ridge regression \citep{van2011mice}, and MissForest \citep{stekhoven2012missforest}. 

\begin{figure}[t]
% \vspace{-3mm}
\centering
\subfigure[]{%
	\begin{minipage}[b]{0.47\linewidth}%
		\includegraphics[width=1\textwidth]{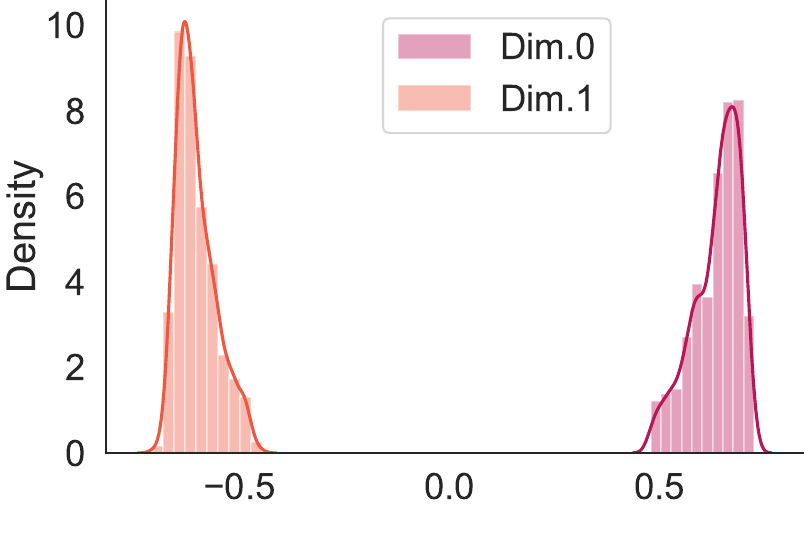}%
	\end{minipage}%
	
}%
\subfigure[]{%
	\begin{minipage}[b]{0.47\linewidth}%
		\includegraphics[width=\textwidth]{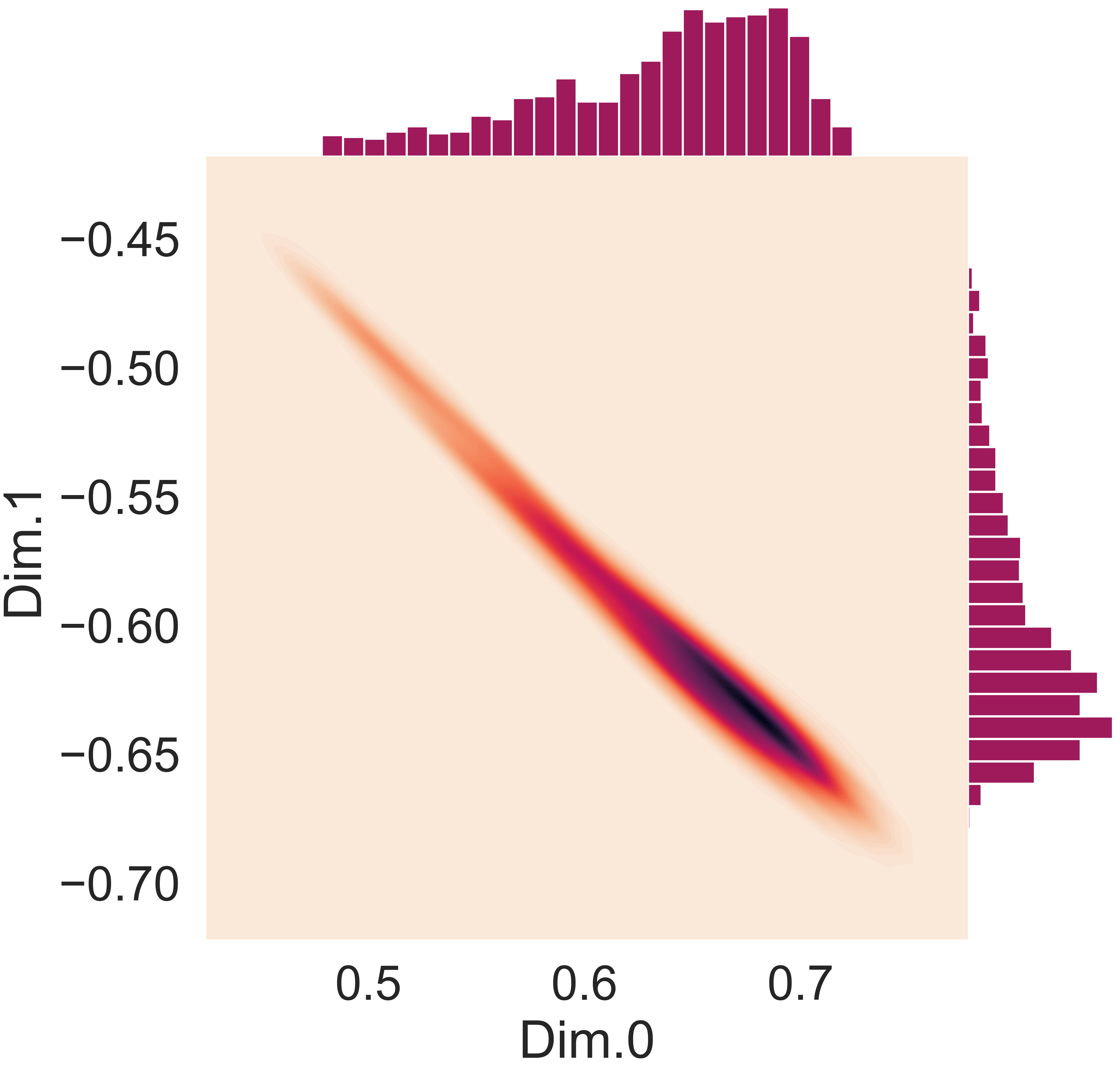}%
	\end{minipage}%
}%
\vspace{-3mm}
\caption{Visualization of latent space. (a): the value distribution of latent space after dimension reduction; (b): the smooth kernel density of the two dimensions in (a).}
\label{latent visualization}
\vspace{-2mm}
\end{figure}

In Table \ref{fig gauss}, we first notice that MIWAE performs poorly on MNAR data, which indicates that a proper model of the missing mechanism is necessary for tackling the MNAR problem. Faced with multiple missing scenarios and mixed missing scenarios, our proposed GNR surpasses baselines with significant margins consistently (averagely improved by 13.3\%), which verifies the validity of the proposed conjunction model for deep generative imputation models.

\subsubsection{\textbf{Visualization of latent space.}} The vanilla implementation of the VAE assumes a standard Gaussian marginal (prior) over latent variable $\mathbf{z}$. The latent variables (or aggregated posterior variables) are approximately isotropic standard Gaussian in each dimension. However, after the dimension reduction of latent variables in GNR, we generate two separate populations of posterior distributions with significantly different ranges in Figure~\ref{latent visualization}(a). We argue that GNR couples the missing mechanism into latent representation, the two peaks in Figure~\ref{latent visualization}(a) focus on missing values and observed values respectively. Each sample contains both observed and missing values, thus each sample has dual attributes responding to the two separate populations of posterior distributions. The kernel density in Figure~\ref{latent visualization}(b) is not an ordinary radial circle representing each dimension independently and has structured information. We hypnosis that to prevent all the values in a sample fall into the missing-data modeling or observed-data modeling, the latent variables in different dimensions tend to separate.

\subsection{Single imputation on UCI datasets}
\label{UCI}

\quad We compare different imputation methods on real-world datasets from the UCI repository \citep{asuncion2007uci} \footnote{\url{https://archive-beta.ics.uci.edu/}}. 
% Statistics of the datasets see Table~\ref{statistics} in Appendix.
Similar to \citet{ipsen2020not}, we set the dimension of the latent space to $d-1$ ($\mathbf{x} \in \mathbb{R}^\text{d}$), and use zero imputation and standardization for each feature before the missing is introduced. Experiments are repeated 5 times. 

The interrelationships between features are more complex in real-world datasets compared to synthetic datasets. GNR in Table~\ref{fig UCI} shows significant margins over other baselines clearly in multiple datasets (9.9\%, 10.3\%, and 18.8\% respectively for 20\%, 80\%, and 100\% missing probability). Similar to results in \citet{ghalebikesabi2021deep}, traditional methods such as MissForest and MICE sometimes outperform neural network-based methods when the missing probability is low. But when we sample a single value from the predictive distribution of traditional methods, deep generative models outperform them. Another drawback of MICE and MissForest is that they are typically not scalable to high-dimensional datasets, while GNR can. The stable and superior performance compared to the baselines in different datasets with multiple missing probabilities demonstrates the broad applicability and robustness of our approach.

\begin{figure*}[t]
\vspace{-4mm}%
\centering%
\subfigure{%
\centering%
\begin{minipage}[t]{0.02\linewidth}%
\centering%
\rotatebox{90}{\textbf{\quad Ground Truth}}%
\end{minipage}%
}%
\vspace{-4mm}%
% \hspace{-2mm}%
\centering%
\subfigure{%
\centering%
\begin{minipage}[t]{0.14\linewidth}%
% \textbf{Ground Truth}
\centering%
\includegraphics[width=\textwidth]{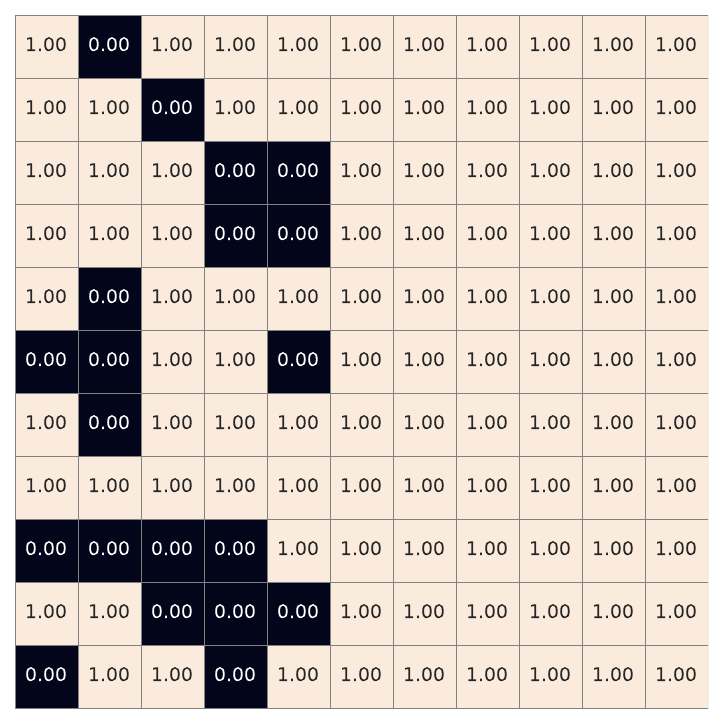}%
\end{minipage}%
}%
% \hspace{-2mm}
\centering%
\subfigure{%
\begin{minipage}[t]{0.14\linewidth}%
\centering%
\includegraphics[width=\textwidth]{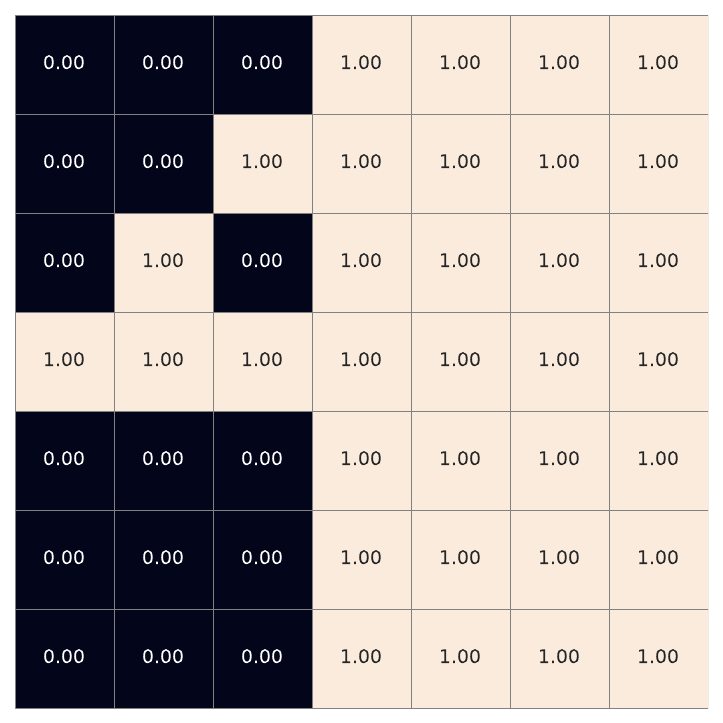}%
\end{minipage}%
}%
% \hspace{-2mm}
\centering%
\subfigure{%
\begin{minipage}[t]{0.14\linewidth}%
\centering%
\includegraphics[width=\textwidth]{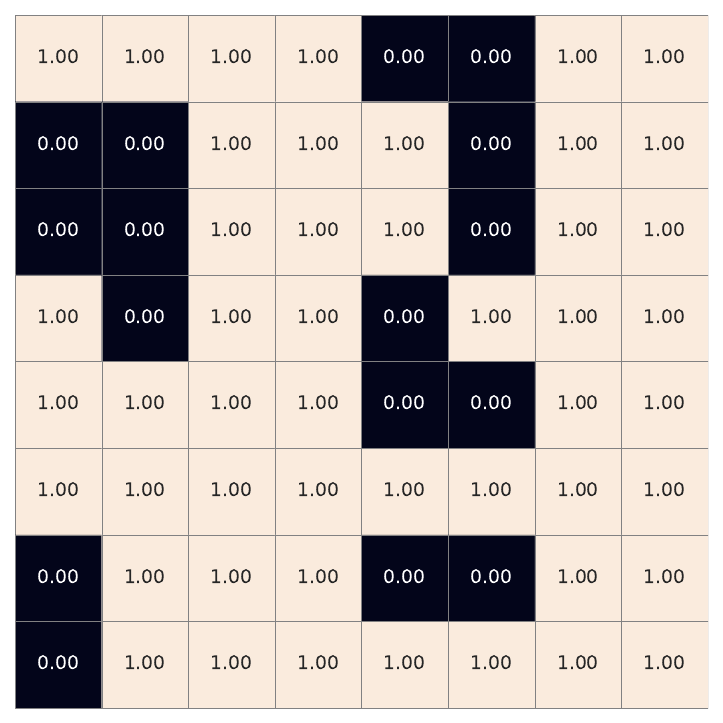}%
\end{minipage}%
}%
% \hspace{-2mm}
\centering%
\subfigure{%
\begin{minipage}[b]{0.14\linewidth}%
\centering%
\includegraphics[width=\textwidth]{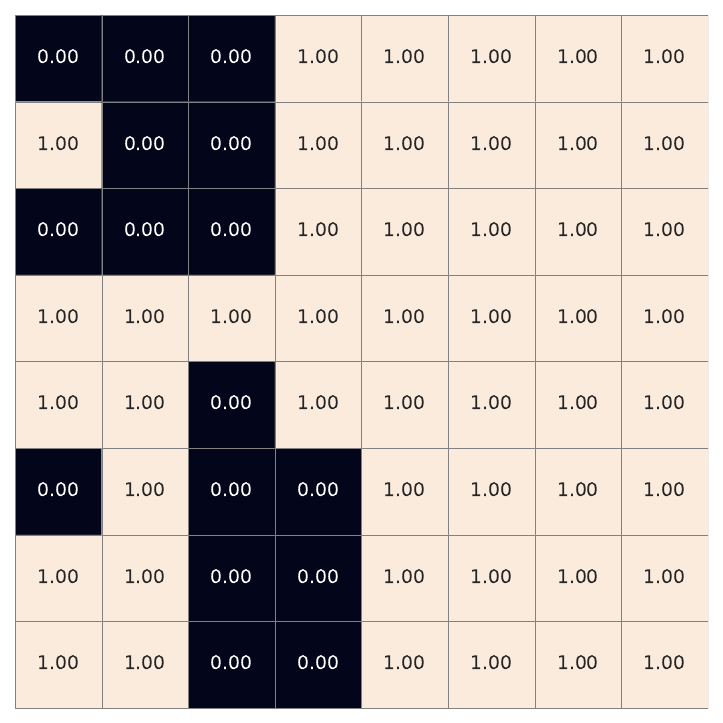}%
\end{minipage}%
}%
% \hspace{-2mm}
\centering%
\subfigure{%
\begin{minipage}[b]{0.14\linewidth}%
\centering%
\includegraphics[width=\textwidth]{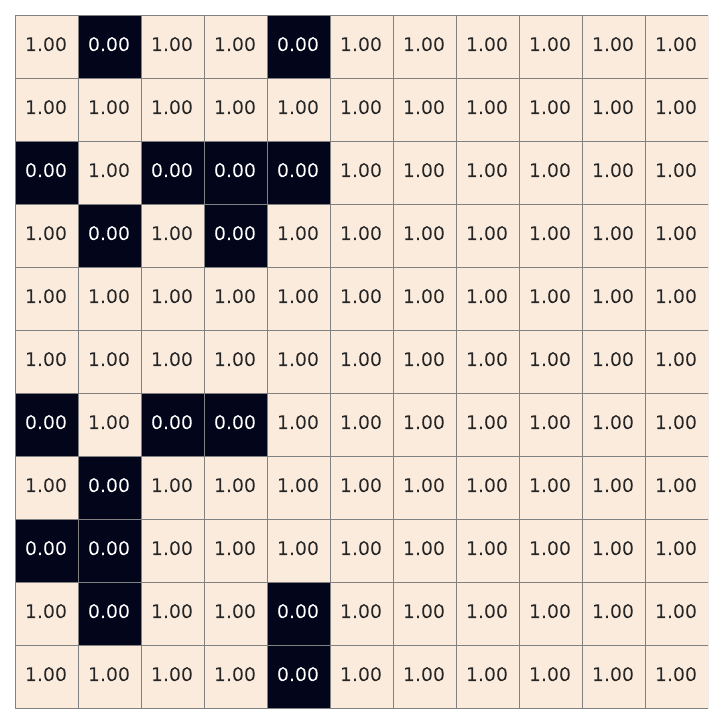}%
\end{minipage}%
}%
% \hspace{-2mm}
\centering%
\subfigure{%
\begin{minipage}[b]{0.14\linewidth}%
\centering%
\includegraphics[width=\textwidth]{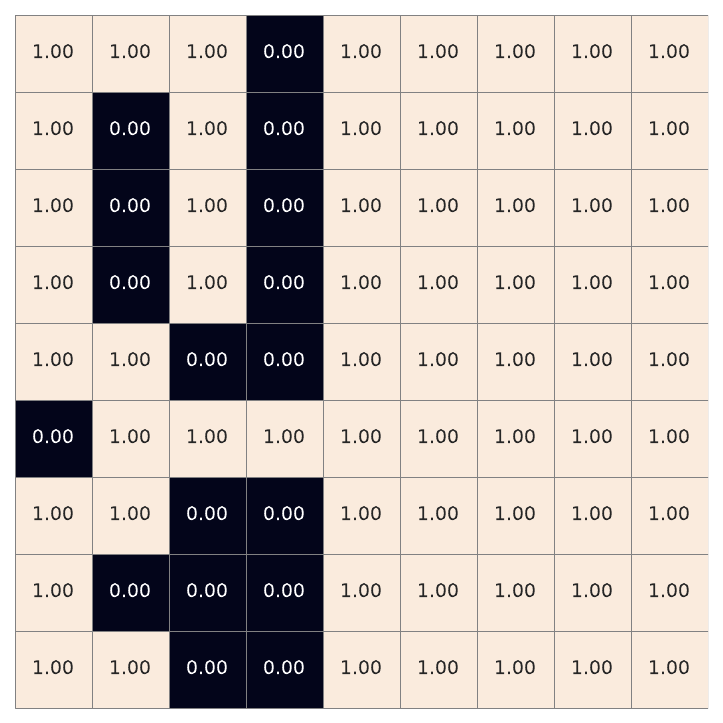}% 
\end{minipage}%
}%
% \hspace{-2mm}
\centering%
\subfigure{%
\begin{minipage}[b]{0.14\linewidth}%
\centering%
\includegraphics[width=\textwidth]{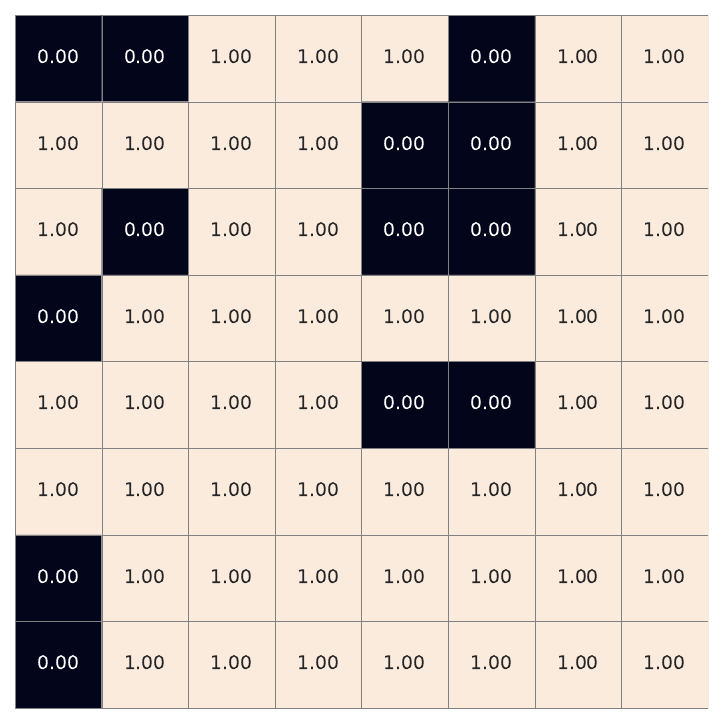}%
\end{minipage}%
}%
% \hspace{-2mm}

\centering%
\subfigure{%
\centering%
\begin{minipage}[t]{0.02\linewidth}%
\centering%
\rotatebox{90}{\textbf{\quad \ not-MIWAE}}%   
\end{minipage}%
}%
% \hspace{-2mm}%
\vspace{-4mm}%
\centering%
\subfigure{%
\begin{minipage}[b]{0.14\linewidth}%
\centering%
\includegraphics[width=\textwidth]{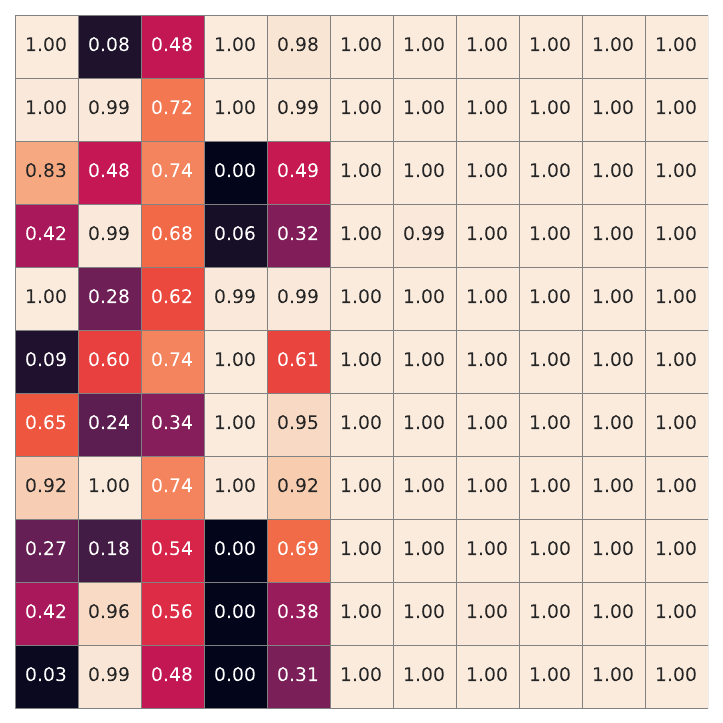}%
\end{minipage}%
}% 
% \hspace{-2mm}
\centering%
\subfigure{%
\begin{minipage}[b]{0.14\linewidth}%
\centering%
\includegraphics[width=\textwidth]{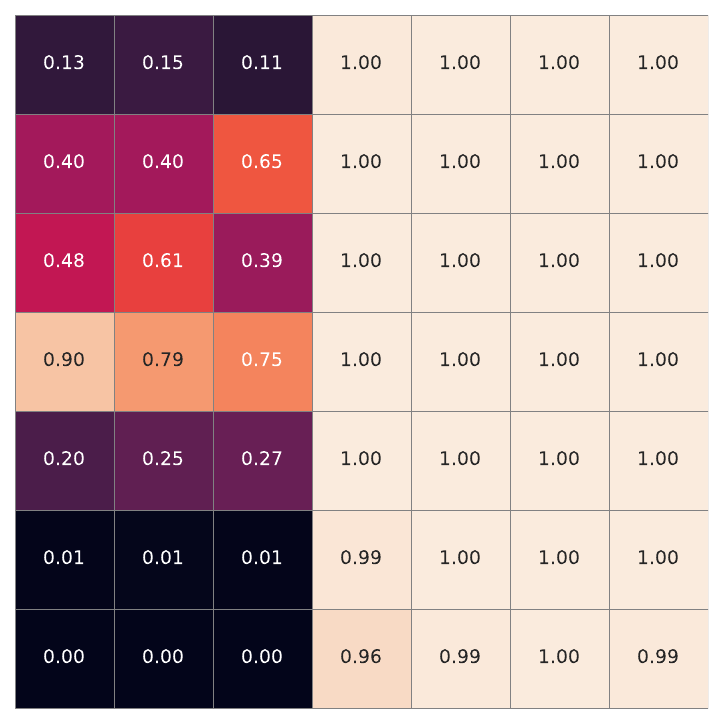}%
\end{minipage}%
}% 
% \hspace{-2mm}
\centering%
\subfigure{%
\begin{minipage}[t]{0.14\linewidth}%
\centering%
\includegraphics[width=1\textwidth]{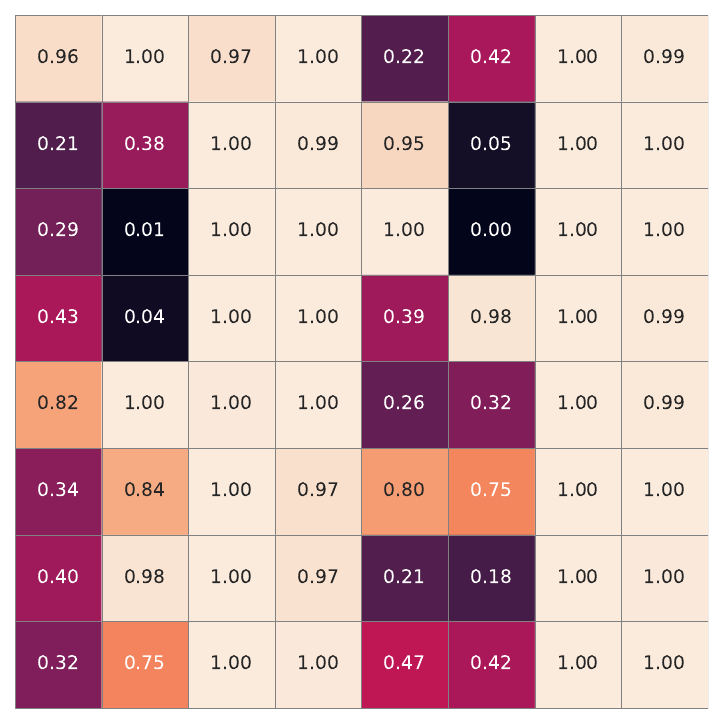}%
\end{minipage}%
}%
% \hspace{-2mm}
\centering%
\subfigure{%
\begin{minipage}[b]{0.14\linewidth}%
\centering
\includegraphics[width=\textwidth]{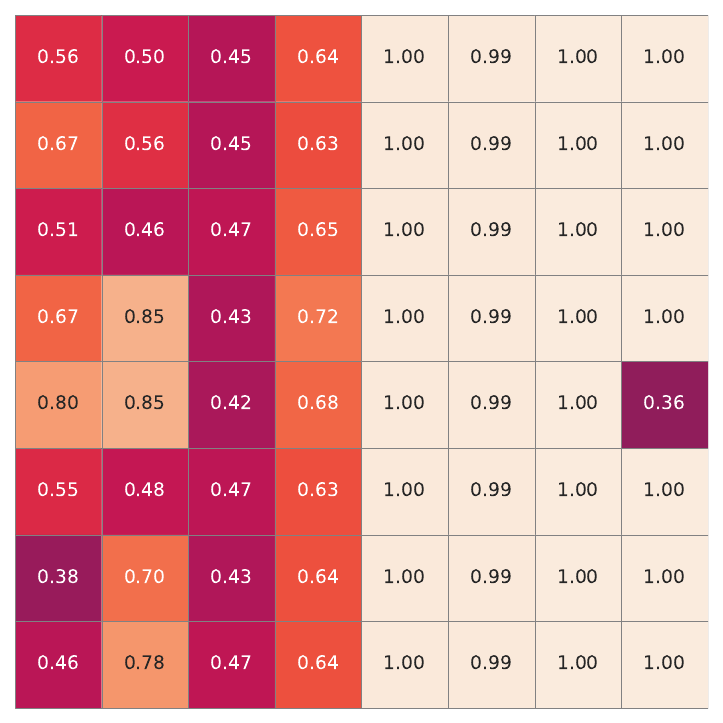}%
\end{minipage}%
}%
% \hspace{-2mm}
\centering%
\subfigure{%
\begin{minipage}[b]{0.14\linewidth}%
\centering%
\includegraphics[width=\textwidth]{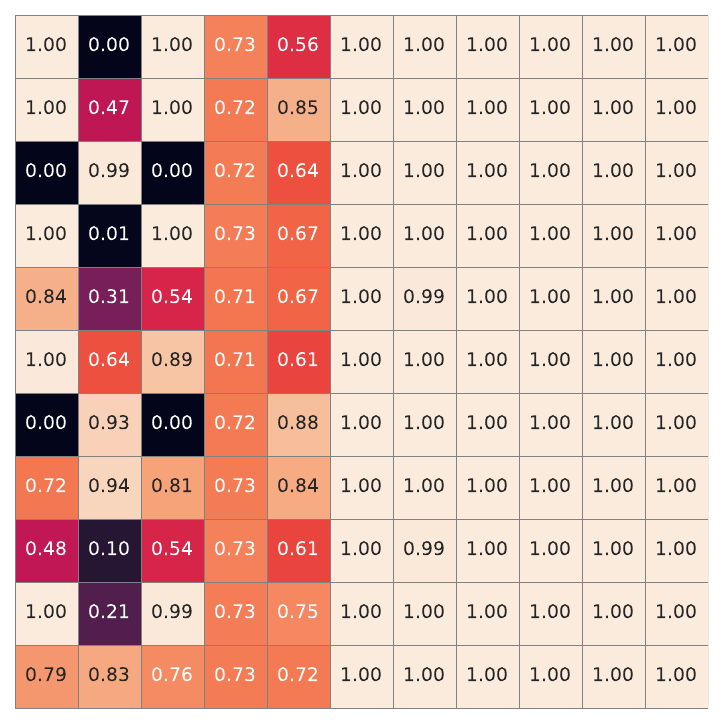}%
\end{minipage}%
}% 
% \hspace{-2mm}
\centering%
\subfigure{%
\begin{minipage}[b]{0.14\linewidth}%
\centering%
\includegraphics[width=\textwidth]{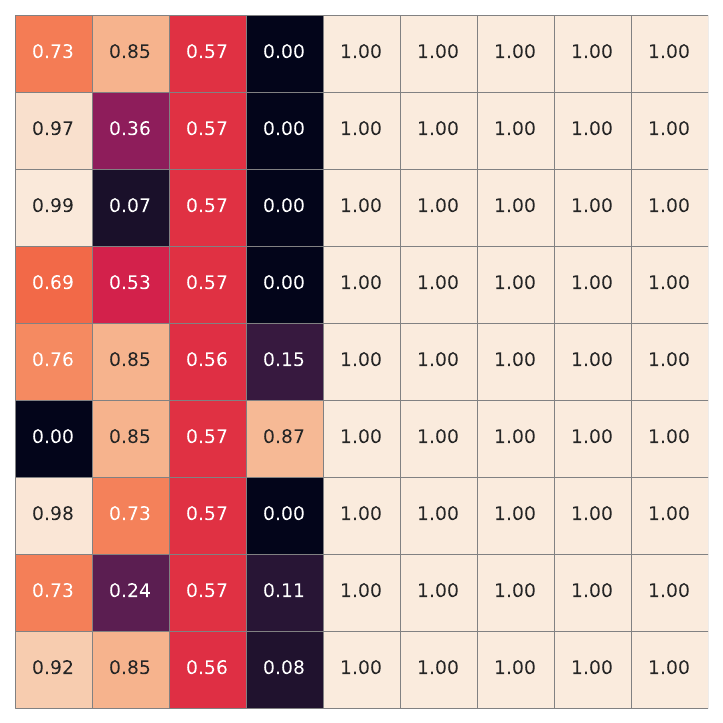}%
\end{minipage}%
}% 
% \hspace{-2mm}
\centering%
\subfigure{%
\begin{minipage}[b]{0.14\linewidth}%
\centering%
\includegraphics[width=\textwidth]{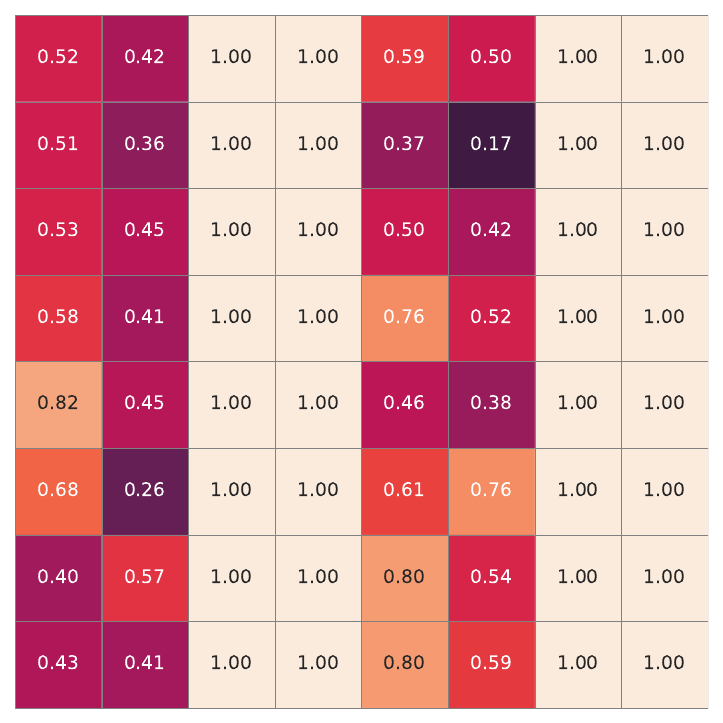}%
\end{minipage}%
}%
% \hspace{-2mm}

% \hspace{-2mm}
\centering%
\subfigure{%
\centering%
\begin{minipage}[t]{0.02\linewidth}%
\centering%
\rotatebox{90}{\textbf{\qquad \ \; GNR}}%
\end{minipage}%
}%
\setcounter{subfigure}{0}%
% \vspace{-2mm}%
% \hspace{-2mm}%
\centering%
\subfigure[\textbf{\textit{white wine}}]{%
\begin{minipage}[b]{0.14\linewidth}%
\centering%
\includegraphics[width=\textwidth]{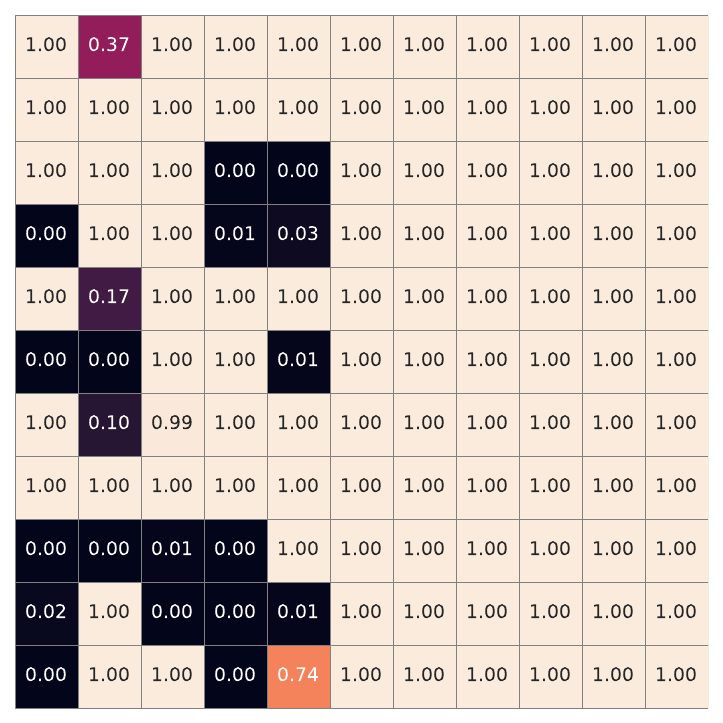}%
\end{minipage}%
}% 
% \hspace{-2mm}
\centering%
\subfigure[\textbf{\textit{abalone}}]{%
\begin{minipage}[t]{0.14\linewidth}%
\centering%
\includegraphics[width=1\textwidth]{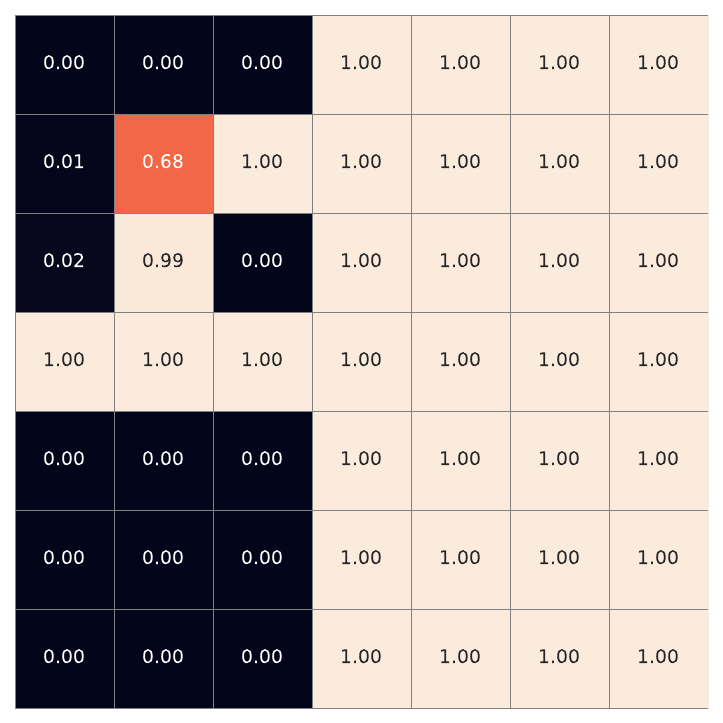}%
\end{minipage}%
}% 
% \hspace{-2mm}
\centering%
\subfigure[\textbf{\textit{banknote}}]{%
\begin{minipage}[t]{0.14\linewidth}%
\centering%
\includegraphics[width=1\textwidth]{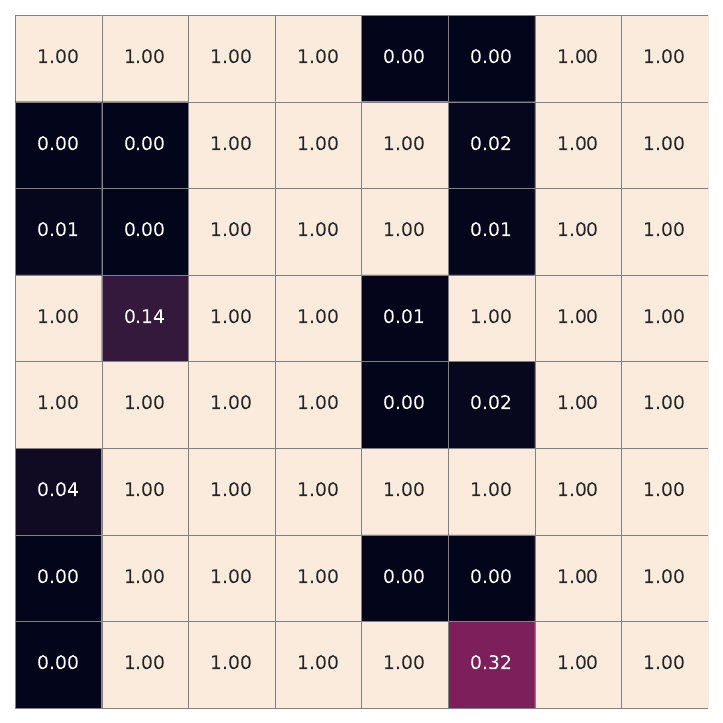}%
\end{minipage}%
}% 
% \hspace{-2mm}
\centering%
\subfigure[\textbf{\textit{yeast}}]{%
\begin{minipage}[b]{0.14\linewidth}%
\centering%
\includegraphics[width=\textwidth]{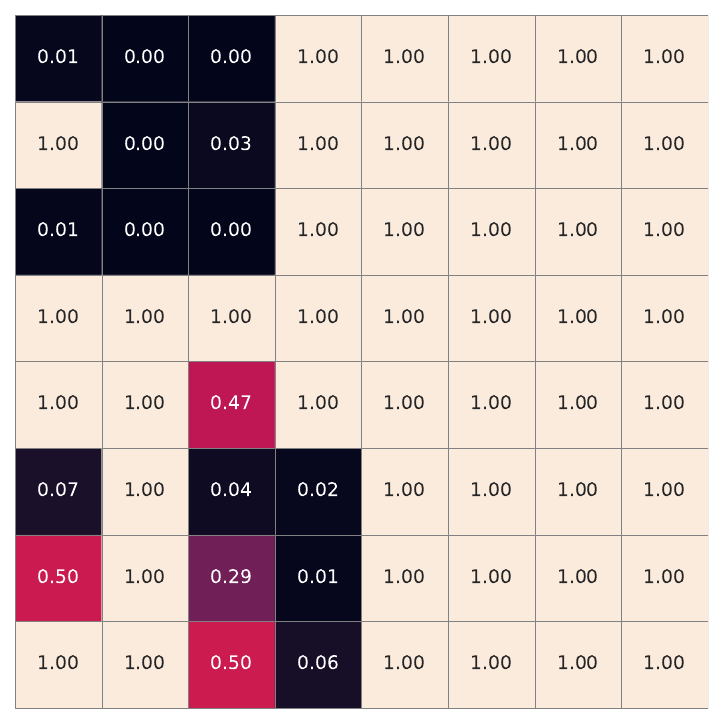}%
\end{minipage}%
}% 
% \hspace{-2mm}
\centering%
\subfigure[\textbf{\textit{red wine}}]{%
\begin{minipage}[b]{0.14\linewidth}%
\centering%
\includegraphics[width=\textwidth]{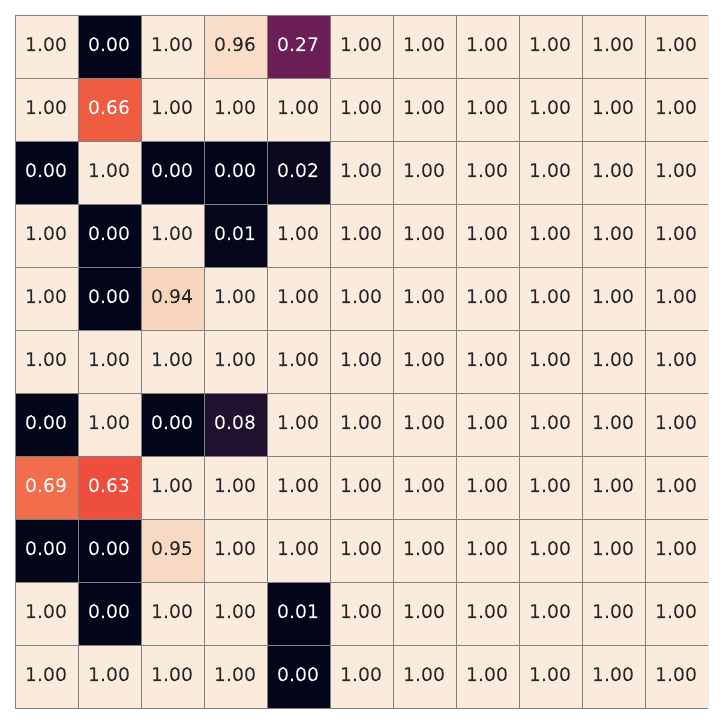}%
\end{minipage}%
}% 
% \hspace{-2mm}
\centering%
\subfigure[\textbf{\textit{concrete}}]{%
\begin{minipage}[b]{0.14\linewidth}%
\centering%
\includegraphics[width=\textwidth]{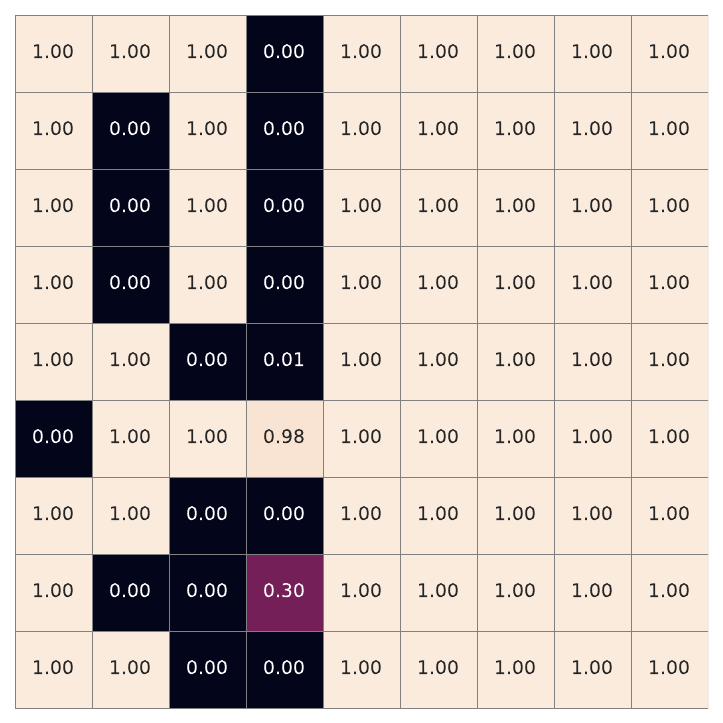}%
\end{minipage}%
}%
% \hspace{-2mm}
\centering%
\subfigure[\textbf{\textit{Gaussian}}]{%
\begin{minipage}[b]{0.14\linewidth}%
\centering%
\includegraphics[width=\textwidth]{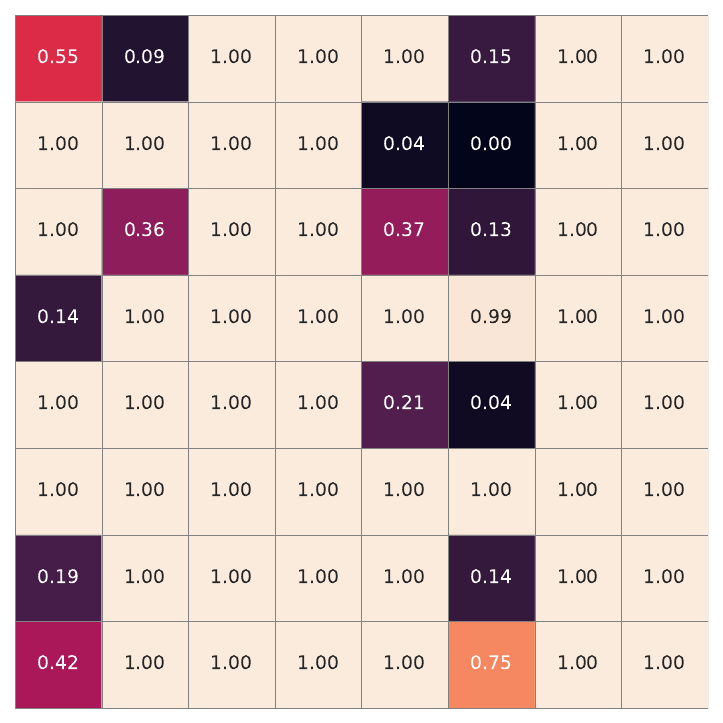}%
\end{minipage}%
}%
% \hspace{-2mm}
% \hspace{-2mm}
\vspace{-2mm}
\caption{Random local details of the reconstructed undiscretized mask in 6 realistic datasets and 1 synthetic dataset. The value in the undiscretized mask can be considered as the probability that a value is observed, namely \textit{probabilistic mask}.}%
\label{mask}%
% \vspace{-2mm}
\end{figure*}

\begin{figure*}[t]
\vspace{-2mm}
	\centering     
	\includegraphics[width=\linewidth,trim=0 5 0 5,clip]{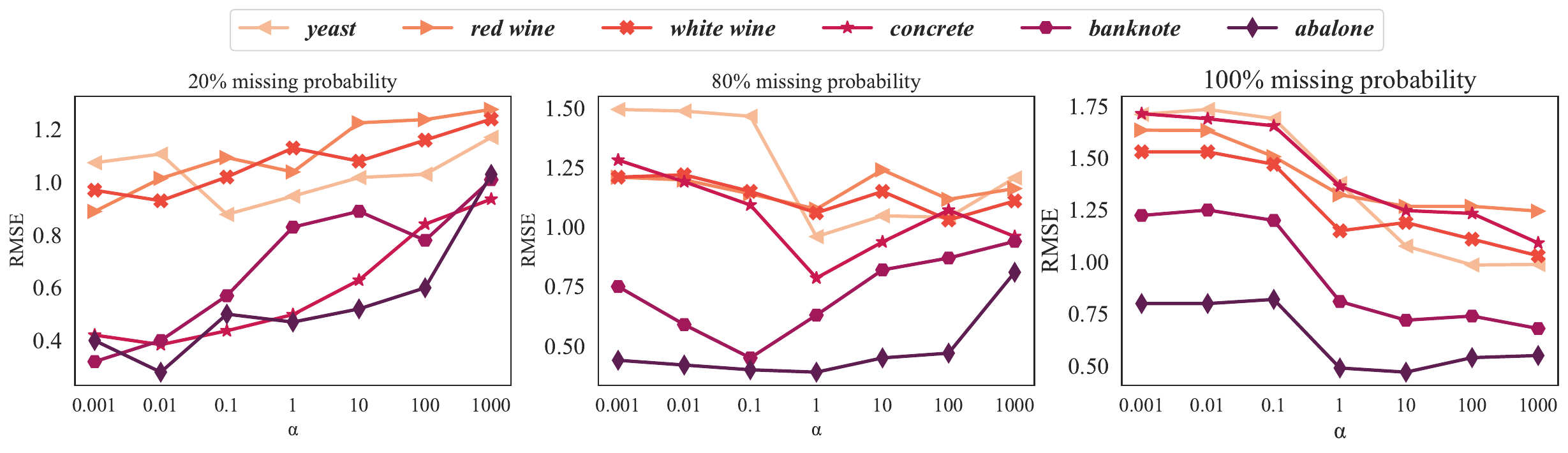} 
  \vspace{-6mm}
	%trim=左 下 右 上
	\caption{Performance comparisons with varying hyper-parameter values $\alpha$ on UCI datasets.}    
% \vspace{-2mm}
	\label{alpha} 
\end{figure*}

\subsubsection{\textbf{Studies of missing mask reconstruction.}}
We also demonstrate the mask-reconstruction accuracy in Table~\ref{mask_acc}. Firstly we assume the MCAR scenario that observed data and missing data have the same distribution and randomly generate the reconstructed mask. To get the best accuracy, $\sum p(\mathbf{m}=0)=k/4$, and $\sum p(\mathbf{m}=1)=1-k/4$, must be satisfied (where $k$ is the missing probability). In this case, the theoretical accuracy is $[k^2+(1-k/2)^2]*100\%$, which is the lower bound of reconstructed mask accuracy. One of our core baselines, PSMVAE, is not compared because it uses both $\mathbf{x}$ and $\mathbf{m}$ as input. It is hard to figure out whether PSMVAE has learned the missing mechanism to build the missing mask or PSMVAE just recovers the inputted $\mathbf{m}$. 

In Table~\ref{mask_acc}, not-MIWAE usually gives a poor mask-reconstruction performance and is sometimes close to the lower bound (RANDOM), which means not-MIWAE fails to extract the information in the mask. The visualization in Figure~\ref{mask}(a) proves our point: not-MIWAE pushes some values, which are difficult to determine whether missing, to 0.5 (close to red) to reduce the global loss. GNR (Figure~\ref{mask}(b)) gives a convincing probabilistic mask that is similar to the ground truth (Figure~\ref{mask}(c)), which implies that GNR understands the missing mechanism from its learning process and makes the imputation task more principled. 
Taking Table~\ref{fig UCI} and Table~\ref{mask_acc} together, the parallel-structure GNR fully mines the information in incomplete data and missing masks, and finally performs better than others. Note that even in some cases our GNR does not significantly outperform other baselines on global evaluation metrics, GNR still gives more reasonable estimates for each local missing value based on accurate mask reconstruction, and principled imputation results are more meaningful for downstream tasks.

\subsubsection{\textbf{Studies of hyper-parameter.}} As a proportionality coefficient, the optimal choice of $\alpha$ only depends on the specific missing scenario of the dataset instead of the network architecture of the practical algorithm. In Figure~\ref{alpha},
datasets have different sensitivities to $\alpha$ because of complex interrelationships between their features. But we notice that for each dataset, the optimal choice of $\alpha$ is positively correlated with the missing probability. $\alpha$ balances a trade-off between the flexibility of the missing mask model and the distortion it induces in the data model when the data is MAR/MCAR. This can be explained by the fact that GNR needs to spend more effort to reconstruct a more realistic missing mask and limit the flexibility of the missing mask model when the missing probability is high. On the contrary, in extreme cases where the MNAR mechanism induces little distortion to the data distribution, we do not even need a missing mask model ($\alpha\to0$) and degenerate GNR to a model for MAR/MCAR data similar to MIWAE \citep{mattei2019miwae}. 
Note that for different application scenarios, we do not need a more detailed range of $\alpha$, and directly select an approximate $\alpha$ from \{0.01, 1, 100\} according to the sparsity of the dataset.

\subsection{Imputation with MCAR test sets}
\label{RCT} 

\quad We employ two datasets, \textit{Yahoo!R3} and \textit{Coat} (Figure~\ref{data distribution}), which contain an MNAR set and a small MCAR set:\\
% \begin{itemize}[leftmargin=*]   
$\bullet$\textit{Yahoo!R3} contains five-star user-song ratings. The MNAR training set contains more than 300K self-selected ratings from 15,400 users on 1,000 songs (with 0.94 sparsity). And the MCAR test set contains randomly selected ratings from 5,400 users on 10 random songs (with 0.99 sparsity). \\
$\bullet$\textit{Coat} contains five-star user-coat ratings from 290 Amazon Mechanical Turk workers on an inventory of 300 coats. The training set contains 6,500 MNAR ratings collected through self-selections by the Turk workers (with 0.92 sparsity). And the test set is MCAR collected by asking the Turk workers to rate 16 randomly selected coats (with 0.95 sparsity).\\
% \end{itemize}
We train GNR and baselines on MNAR training sets and evaluate the imputation error on MAR test sets to get an unbiased estimate of the imputation error. 
We compare with more baselines:\\
% \begin{itemize}[leftmargin=*]
	$\bullet$ MF \citep{koren2009matrix}: the basic matrix factorization model makes no further assumptions about the missing mechanism;\\
	$\bullet$ MF-IPS and MF-SNIPS \citep{schnabel2016recommendations}: each data uses the inverse of its propensity score to weight for unbiased performance estimation;\\
	$\bullet$ MF-DR and MF-DR-JL \citep{wang2019doubly}: combining the propensity-scoring approach \citep{schnabel2016recommendations} with an error-imputation approach by \citet{steck2013evaluation} to obtain a doubly robust estimator;\\
	$\bullet$ MF-CVIB \citep{wang2020information}: a counterfactual variational information bottleneck used for debiasing learning without MAR data;\\
	$\bullet$ GAIN \citep{yoon2018gain}: based on GANs, in which the generator outputs imputed data, and the discriminator determines which variables are observed based on partial information of the missing mask.
	% \end{itemize}

The propensity-based methods (MF-IPS, MF-SNIPS, MF-DR-JL) need access to a small sample of MAR data to estimate the propensities, which use 5\% of the MAR test set for training. 
Similar to \citet{ipsen2020not}, we use the permutation invariant encoder \cite{ma2018eddi} with an embedding size of 20 and a code size of 50, along with a linear mapping to a latent space of size 30. We use Equation~\ref{data} to preprocess the data, where $r \in \{1, 2, 3, 4, 5\}$ denotes a five-star rating. Additionally, $\epsilon$ controls the noise level in the grade information. We apply $\epsilon \sim N(0,0.1)$ for training sets and $\epsilon = 0 $ for test sets. The output layer has a sigmoid activation for the mean in the data model, scaled to match the scale of the inputs.
 \begin{equation}
	\hat{r} = \epsilon+(1-\epsilon) \frac{2^{r}-1}{2^{r_{\max }}-1}.
	\label{data}
\end{equation}

% \begin{table}[t]    
% 	\centering   
%  \caption{Test MSE on \textit{Yahoo!R3} and \textit{Coat}.}  \label{fig RCT}    
%  \begin{tabular}{lcc}      
%  \toprule[1.25pt]    
%  \makebox[0.3\linewidth][l]{\textbf{Method}} 
%  & \makebox[0.2\linewidth][c]{\textit{\textbf{Yahoo!R3}}} & \makebox[0.2\linewidth][c]{\textit{\textbf{Coat}}} 
%  \\      \midrule[0.75pt]      
%  \hdashline
%  \multicolumn{3}{l}{\textit{VAE-based methods}} 
%  \\ \hdashline MIWAE & 1.16  & 1.38 
%  \\      not-MIWAE & \underline{1.08}  & 1.29 
%  \\      PSMVAE  &  1.13   & \underline{1.21}
%  \\     \textbf{GNR} & \textbf{0.95}  & \textbf{1.11} 
%  \\   \hdashline     \multicolumn{3}{l}{\textit{MF-based methods}} 
%  \\ \hdashline MF    & 1.26  & 1.28  
%  \\      MF-IPS & 1.18  & 1.33 
%  \\      MF-SNIPS & 1.10  & 1.23
%  \\      MF-DR & 1.15  & 1.31 
%  \\   MF-DR-JL &1.09 &1.22 
%  \\  MF-CVIB & 1.32  & 1.24  
%  \\   \hdashline      \multicolumn{3}{l}{\textit{other method}} \\  \hdashline GAIN &1.11 &1.23 
%  \\ \midrule[0.75pt] 
% 		\%Improv. & +12.0\% & +8.3\%
%   \\ \bottomrule[1.25pt]       
%   \end{tabular}      
%   \end{table}

% Please add the following required packages to your document preamble:
% \usepackage{booktabs}
% \usepackage{multirow}
% \usepackage{graphicx}
% Please add the following required packages to your document preamble:
% \usepackage{booktabs}
% \usepackage{multirow}
% \usepackage{graphicx}

Results are shown in Table~\ref{fig RCT}. Our GNR outperforms all the baselines, and improvements are rather impressive --- 12.0\% and 8.3\% in terms of \textit{Yahoo!R3} and \textit{Coat}. These results validate that GNR can better understand the MNAR mechanism and match the distribution of MCAR test sets. MNAR-based models generally outperform their MAR/MCAR versions (MIWAE, MF) for the appropriate assumption about the missing mechanism. And the propensity-based methods suffer from high variance and are difficult to develop a proper propensity score, which damages the performance. More importantly, among VAE-based models which use similar settings, GNR (based on the conjunction model) outperforms its selection model counterpart (not-MIWAE) and pattern-set mixture model counterpart (PSMVAE).

\section{Related work} 
% \quad We first briefly review traditional methods that deal with MCAR and MAR data. \citet{rubin1976inference} introduces and formalizes the appropriateness of ignoring the missing process when doing likelihood-based or Bayesian inference. Then \citet{dempster1977maximum} propose the EM algorithm making it possible to obtain a maximum likelihood estimate to handle missing data. Most traditional imputation methods are limited by the prior assumption that the missing mechanism is ignorable and can be roughly classified into two categories: single imputation methods (SI) and multiple imputation methods (MI). Single imputation only produces one single set of imputed values for each data instance, including mean/zero imputation \citep{nazabal2020handling}, MissForest \citep{stekhoven2012missforest}, and matrix completion \citep{mazumder2010spectral}. As opposed to single imputation, the multiple imputations (MI) methods such as MICE \citep{van2011mice}, first proposed by Rubin \citep{rubin1988overview}, return multiple imputation values for subsequent statistical analysis. Unlike single imputation, the standard errors of estimated parameters produced by MI are unbiased when data are MCAR/MAR. In addition, there exist other methods, such as inverse probability weighting \citep{horton2001multiple}, which can be directly applied to MAR without introducing additional bias. However, to prevent bias, these methods cannot be directly utilized under the MNAR assumption.

\quad We briefly review methods that deal with MNAR data. Inverse propensity score (IPS) \citep{hernandez2014probabilistic}, a counterfactual technique, reweighting the collected data for expectation-unbiased learning, is taken to impute \citep{chen2021autodebias,saito2020asymmetric,wang2020information}. \citet{sportisse2020imputation} and \citet{ma2019missing}, use low-rank models for estimation and imputation in MNAR settings. The causal approach is taken to the imputation task where MNAR is treated as a confounding bias \citep{wang2019blessings,kyono2021miracle}. \citet{muzellec2020missing}, use optimal transport to define a relevant loss for imputation.

Growing interest has been focused on applying deep generative models to missing-data imputation. Deep generative models such as generative adversarial nets (GANs) \citep{goodfellow2014generative} or variational autoencoders (VAEs) \citep{kingma2013auto} have been used for imputation since their introduction. Some methods use an extension of the variational lower bound to handle missing data under the MAR assumption \cite{nazabal2020handling,ma2018eddi,ma2018partial}. \citet{yoon2018gain} and \citet{li2019misgan} use GANs for imputing MCAR data. Diffusion models \citep{song2019generative,ho2020denoising} are also used for time series data imputation \citep{tashiro2021csdi} under MAR assumption. More recently, deep generative models under MNAR have been studied. \citet{gong2021variational} and \citet{ipsen2020not} are based on the selection model. \citet{ma2021identifiable} provide an identifiable deep generative selection model for MNAR data. \citet{collier2020vaes} are based on the pattern mixture model. \citet{ghalebikesabi2021deep} build an idea from the pattern-set mixture model and introduce a probabilistic semi-supervised approach.

\begin{table}[t]
\centering
\renewcommand\arraystretch{1.1}
% \vspace{-2mm}
\caption{Imputation MSE on \textit{Yahoo!R3} and \textit{Coat}.} 
\vspace{-2mm}
\label{fig RCT}
\begin{tabular}{@{}lccc@{}}
\toprule[1.25pt]     
\makebox[0.25\linewidth][l]{\textbf{Method}} &
  \makebox[0.2\linewidth][c]{\textbf{Family}} &
  \makebox[0.2\linewidth][c]{\textit{\textbf{Yahoo!R3}}} &
  \makebox[0.2\linewidth][c]{\textit{\textbf{Coat}}} \\ \midrule

MIWAE   & \multirow{4}{*}{VAE} & 1.16             & 1.38             \\
not-MIWAE          &                      & \underline{1.08} & 1.29             \\
PSMVAE             &                      & 1.13             & \underline{1.21} \\
\textbf{GNR}       &                      & \textbf{0.95}    & \textbf{1.11}    \\
\hdashline MF      & \multirow{6}{*}{MF}  & 1.26             & 1.28             \\
MF-IPS             &                      & 1.18             & 1.33             \\
MF-SNIPS           &                      & 1.10             & 1.23             \\
MF-DR              &                      & 1.15             & 1.31             \\
MF-DR-JL           &                      & 1.09             & 1.22             \\
MF-CVIB            &                      & 1.32             & 1.24             \\
\hdashline GAIN    & GAN                  & 1.11             & 1.23             \\

\midrule[0.75pt]
\%Improv. &                     & +12.0\%          & +8.3\%           \\ \bottomrule
\end{tabular}%
\vspace{-2mm}
\end{table}

\section{Conclusion} 
\quad In this work, we propose principled solutions from the perspectives of modeling the MNAR mechanism and application area. First, we exemplary explain some statistical MNAR modeling approaches and point out the inherent flaws in existing deep generative imputation models. Second, we propose a new MNAR modeling approach, the conjunction model, which views the complete data and the missing mask as two modalities of missing-data multimodality from a novel multimodal perspective. Third, we propose a practical algorithm, GNR, which fully mines the information in the two modalities without interfering with each other. The visualization of the reconstructed mask and latent variable, and the study of the hyper-parameter provide interpretability to GNR. Finally, we demonstrate that our method gains a significant and robust improvement over existing baselines on imputation tasks under various missing settings for both synthetic and real-world datasets. 

For future work, we are interested in several extensions: (i) extending GNR to deep supervised learning; (ii) introducing identifiability and causal discovery to GNR \citep{khemakhem2020variational}; (iii) modeling the conjunction model based on the deep diffusion models\citep{ho2020denoising,song2020score}.

%%
%% The acknowledgments section is defined using the "acks" environment
%% (and NOT an unnumbered section). This ensures the proper
%% identification of the section in the article metadata, and the
%% consistent spelling of the heading.
% \begin{acks}
	% To Robert, for the bagels and for explaining CMYK and color spaces.
	% \end{acks}
%%
%% The next two lines define the bibliography style to be used, and
%% the bibliography file.
\bibliographystyle{ACM-Reference-Format}
\balance
\bibliography{sample-base}

%%% -*-BibTeX-*-
%%% Do NOT edit. File created by BibTeX with style
%%% ACM-Reference-Format-Journals [18-Jan-2012].

\begin{thebibliography}{40}

%%% ====================================================================
%%% NOTE TO THE USER: you can override these defaults by providing
%%% customized versions of any of these macros before the \bibliography
%%% command.  Each of them MUST provide its own final punctuation,
%%% except for \shownote{}, \showDOI{}, and \showURL{}.  The latter two
%%% do not use final punctuation, in order to avoid confusing it with
%%% the Web address.
%%%
%%% To suppress output of a particular field, define its macro to expand
%%% to an empty string, or better, \unskip, like this:
%%%
%%% \newcommand{\showDOI}[1]{\unskip}   % LaTeX syntax
%%%
%%% \def \showDOI #1{\unskip}           % plain TeX syntax
%%%
%%% ====================================================================

\ifx \showCODEN    \undefined \def \showCODEN     #1{\unskip}     \fi
\ifx \showDOI      \undefined \def \showDOI       #1{#1}\fi
\ifx \showISBNx    \undefined \def \showISBNx     #1{\unskip}     \fi
\ifx \showISBNxiii \undefined \def \showISBNxiii  #1{\unskip}     \fi
\ifx \showISSN     \undefined \def \showISSN      #1{\unskip}     \fi
\ifx \showLCCN     \undefined \def \showLCCN      #1{\unskip}     \fi
\ifx \shownote     \undefined \def \shownote      #1{#1}          \fi
\ifx \showarticletitle \undefined \def \showarticletitle #1{#1}   \fi
\ifx \showURL      \undefined \def \showURL       {\relax}        \fi
% The following commands are used for tagged output and should be
% invisible to TeX
\providecommand\bibfield[2]{#2}
\providecommand\bibinfo[2]{#2}
\providecommand\natexlab[1]{#1}
\providecommand\showeprint[2][]{arXiv:#2}

\bibitem[Asuncion and Newman(2007)]%
        {asuncion2007uci}
\bibfield{author}{\bibinfo{person}{Arthur Asuncion} {and}
  \bibinfo{person}{David Newman}.} \bibinfo{year}{2007}\natexlab{}.
\newblock \bibinfo{title}{UCI machine learning repository}.
\newblock
\newblock


\bibitem[Burda et~al\mbox{.}(2016)]%
        {burda2015importance}
\bibfield{author}{\bibinfo{person}{Yuri Burda}, \bibinfo{person}{Roger~B.
  Grosse}, {and} \bibinfo{person}{Ruslan Salakhutdinov}.}
  \bibinfo{year}{2016}\natexlab{}.
\newblock \showarticletitle{Importance Weighted Autoencoders}. In
  \bibinfo{booktitle}{\emph{4th International Conference on Learning
  Representations, {ICLR} 2016, San Juan, Puerto Rico, May 2-4, 2016,
  Conference Track Proceedings}}.
\newblock


\bibitem[Chen et~al\mbox{.}(2021)]%
        {chen2021autodebias}
\bibfield{author}{\bibinfo{person}{Jiawei Chen}, \bibinfo{person}{Hande Dong},
  \bibinfo{person}{Yang Qiu}, \bibinfo{person}{Xiangnan He},
  \bibinfo{person}{Xin Xin}, \bibinfo{person}{Liang Chen},
  \bibinfo{person}{Guli Lin}, {and} \bibinfo{person}{Keping Yang}.}
  \bibinfo{year}{2021}\natexlab{}.
\newblock \showarticletitle{AutoDebias: Learning to debias for recommendation}.
  In \bibinfo{booktitle}{\emph{Proceedings of the 44th International ACM SIGIR
  Conference on Research and Development in Information Retrieval}}.
\newblock


\bibitem[Chen et~al\mbox{.}(2023)]%
        {chen2023bias}
\bibfield{author}{\bibinfo{person}{Jiawei Chen}, \bibinfo{person}{Hande Dong},
  \bibinfo{person}{Xiang Wang}, \bibinfo{person}{Fuli Feng},
  \bibinfo{person}{Meng Wang}, {and} \bibinfo{person}{Xiangnan He}.}
  \bibinfo{year}{2023}\natexlab{}.
\newblock \showarticletitle{Bias and debias in recommender system: A survey and
  future directions}.
\newblock \bibinfo{journal}{\emph{ACM Transactions on Information Systems}}
  \bibinfo{volume}{41}, \bibinfo{number}{3} (\bibinfo{year}{2023}),
  \bibinfo{pages}{1--39}.
\newblock


\bibitem[Collier et~al\mbox{.}(2020)]%
        {collier2020vaes}
\bibfield{author}{\bibinfo{person}{Mark Collier}, \bibinfo{person}{Alfredo
  Nazabal}, {and} \bibinfo{person}{Christopher~KI Williams}.}
  \bibinfo{year}{2020}\natexlab{}.
\newblock \showarticletitle{VAEs in the presence of missing data}.
\newblock \bibinfo{journal}{\emph{arXiv preprint arXiv:2006.05301}}
  (\bibinfo{year}{2020}).
\newblock


\bibitem[Ghalebikesabi et~al\mbox{.}(2021)]%
        {ghalebikesabi2021deep}
\bibfield{author}{\bibinfo{person}{Sahra Ghalebikesabi}, \bibinfo{person}{Rob
  Cornish}, \bibinfo{person}{Chris Holmes}, {and} \bibinfo{person}{Luke~J.
  Kelly}.} \bibinfo{year}{2021}\natexlab{}.
\newblock \showarticletitle{Deep Generative Missingness Pattern-Set Mixture
  Models}. In \bibinfo{booktitle}{\emph{The 24th International Conference on
  Artificial Intelligence and Statistics, {AISTATS} 2021, April 13-15, 2021,
  Virtual Event}} \emph{(\bibinfo{series}{Proceedings of Machine Learning
  Research}, Vol.~\bibinfo{volume}{130})}. \bibinfo{publisher}{{PMLR}}.
\newblock


\bibitem[Gong et~al\mbox{.}(2021)]%
        {gong2021variational}
\bibfield{author}{\bibinfo{person}{Yu Gong}, \bibinfo{person}{Hossein
  Hajimirsadeghi}, \bibinfo{person}{Jiawei He}, \bibinfo{person}{Thibaut
  Durand}, {and} \bibinfo{person}{Greg Mori}.} \bibinfo{year}{2021}\natexlab{}.
\newblock \showarticletitle{Variational Selective Autoencoder: Learning from
  Partially-Observed Heterogeneous Data}. In \bibinfo{booktitle}{\emph{The 24th
  International Conference on Artificial Intelligence and Statistics, {AISTATS}
  2021, April 13-15, 2021, Virtual Event}} \emph{(\bibinfo{series}{Proceedings
  of Machine Learning Research}, Vol.~\bibinfo{volume}{130})}.
  \bibinfo{publisher}{{PMLR}}.
\newblock


\bibitem[Goodfellow et~al\mbox{.}(2014)]%
        {goodfellow2014generative}
\bibfield{author}{\bibinfo{person}{Ian~J. Goodfellow}, \bibinfo{person}{Jean
  Pouget{-}Abadie}, \bibinfo{person}{Mehdi Mirza}, \bibinfo{person}{Bing Xu},
  \bibinfo{person}{David Warde{-}Farley}, \bibinfo{person}{Sherjil Ozair},
  \bibinfo{person}{Aaron~C. Courville}, {and} \bibinfo{person}{Yoshua Bengio}.}
  \bibinfo{year}{2014}\natexlab{}.
\newblock \showarticletitle{Generative Adversarial Nets}. In
  \bibinfo{booktitle}{\emph{Advances in Neural Information Processing Systems
  27: Annual Conference on Neural Information Processing Systems 2014, December
  8-13 2014, Montreal, Quebec, Canada}}.
\newblock


\bibitem[Heckman(1979)]%
        {heckman1979sample}
\bibfield{author}{\bibinfo{person}{James~J Heckman}.}
  \bibinfo{year}{1979}\natexlab{}.
\newblock \showarticletitle{Sample selection bias as a specification error}.
\newblock \bibinfo{journal}{\emph{Econometrica: Journal of the econometric
  society}} (\bibinfo{year}{1979}).
\newblock


\bibitem[Hern{\'{a}}ndez{-}Lobato et~al\mbox{.}(2014)]%
        {hernandez2014probabilistic}
\bibfield{author}{\bibinfo{person}{Jos{\'{e}}~Miguel Hern{\'{a}}ndez{-}Lobato},
  \bibinfo{person}{Neil Houlsby}, {and} \bibinfo{person}{Zoubin Ghahramani}.}
  \bibinfo{year}{2014}\natexlab{}.
\newblock \showarticletitle{Probabilistic Matrix Factorization with Non-random
  Missing Data}. In \bibinfo{booktitle}{\emph{Proceedings of the 31th
  International Conference on Machine Learning, {ICML} 2014, Beijing, China,
  21-26 June 2014}} \emph{(\bibinfo{series}{{JMLR} Workshop and Conference
  Proceedings}, Vol.~\bibinfo{volume}{32})}. \bibinfo{publisher}{JMLR.org}.
\newblock


\bibitem[Ho et~al\mbox{.}(2020)]%
        {ho2020denoising}
\bibfield{author}{\bibinfo{person}{Jonathan Ho}, \bibinfo{person}{Ajay Jain},
  {and} \bibinfo{person}{Pieter Abbeel}.} \bibinfo{year}{2020}\natexlab{}.
\newblock \showarticletitle{Denoising Diffusion Probabilistic Models}. In
  \bibinfo{booktitle}{\emph{Advances in Neural Information Processing Systems
  33: Annual Conference on Neural Information Processing Systems 2020, NeurIPS
  2020, December 6-12, 2020, virtual}}.
\newblock


\bibitem[Ipsen et~al\mbox{.}(2020)]%
        {ipsen2020not}
\bibfield{author}{\bibinfo{person}{Niels~Bruun Ipsen},
  \bibinfo{person}{Pierre-Alexandre Mattei}, {and} \bibinfo{person}{Jes
  Frellsen}.} \bibinfo{year}{2020}\natexlab{}.
\newblock \showarticletitle{not-MIWAE: Deep generative modelling with missing
  not at random data}.
\newblock \bibinfo{journal}{\emph{arXiv preprint arXiv:2006.12871}}
  (\bibinfo{year}{2020}).
\newblock


\bibitem[Khemakhem et~al\mbox{.}(2020)]%
        {khemakhem2020variational}
\bibfield{author}{\bibinfo{person}{Ilyes Khemakhem}, \bibinfo{person}{Diederik
  Kingma}, \bibinfo{person}{Ricardo Monti}, {and} \bibinfo{person}{Aapo
  Hyvarinen}.} \bibinfo{year}{2020}\natexlab{}.
\newblock \showarticletitle{Variational autoencoders and nonlinear ica: A
  unifying framework}. In \bibinfo{booktitle}{\emph{International Conference on
  Artificial Intelligence and Statistics}}. PMLR.
\newblock


\bibitem[Kingma and Welling(2014)]%
        {kingma2013auto}
\bibfield{author}{\bibinfo{person}{Diederik~P. Kingma} {and}
  \bibinfo{person}{Max Welling}.} \bibinfo{year}{2014}\natexlab{}.
\newblock \showarticletitle{Auto-Encoding Variational Bayes}. In
  \bibinfo{booktitle}{\emph{2nd International Conference on Learning
  Representations, {ICLR} 2014, Banff, AB, Canada, April 14-16, 2014,
  Conference Track Proceedings}}.
\newblock


\bibitem[Koren et~al\mbox{.}(2009)]%
        {koren2009matrix}
\bibfield{author}{\bibinfo{person}{Yehuda Koren}, \bibinfo{person}{Robert
  Bell}, {and} \bibinfo{person}{Chris Volinsky}.}
  \bibinfo{year}{2009}\natexlab{}.
\newblock \showarticletitle{Matrix factorization techniques for recommender
  systems}.
\newblock \bibinfo{journal}{\emph{Computer}} \bibinfo{volume}{42},
  \bibinfo{number}{8} (\bibinfo{year}{2009}).
\newblock


\bibitem[Kyono et~al\mbox{.}(2021)]%
        {kyono2021miracle}
\bibfield{author}{\bibinfo{person}{Trent Kyono}, \bibinfo{person}{Yao Zhang},
  \bibinfo{person}{Alexis Bellot}, {and} \bibinfo{person}{Mihaela van~der
  Schaar}.} \bibinfo{year}{2021}\natexlab{}.
\newblock \showarticletitle{MIRACLE: Causally-Aware Imputation via Learning
  Missing Data Mechanisms}.
\newblock \bibinfo{journal}{\emph{Advances in Neural Information Processing
  Systems}}  \bibinfo{volume}{34} (\bibinfo{year}{2021}).
\newblock


\bibitem[Li et~al\mbox{.}(2019)]%
        {li2019misgan}
\bibfield{author}{\bibinfo{person}{Steven~Cheng{-}Xian Li}, \bibinfo{person}{Bo
  Jiang}, {and} \bibinfo{person}{Benjamin~M. Marlin}.}
  \bibinfo{year}{2019}\natexlab{}.
\newblock \showarticletitle{MisGAN: Learning from Incomplete Data with
  Generative Adversarial Networks}. In \bibinfo{booktitle}{\emph{7th
  International Conference on Learning Representations, {ICLR} 2019, New
  Orleans, LA, USA, May 6-9, 2019}}. \bibinfo{publisher}{OpenReview.net}.
\newblock


\bibitem[Little(1993)]%
        {little1993pattern}
\bibfield{author}{\bibinfo{person}{Roderick~JA Little}.}
  \bibinfo{year}{1993}\natexlab{}.
\newblock \showarticletitle{Pattern-mixture models for multivariate incomplete
  data}.
\newblock \bibinfo{journal}{\emph{J. Amer. Statist. Assoc.}}
  \bibinfo{volume}{88}, \bibinfo{number}{421} (\bibinfo{year}{1993}).
\newblock


\bibitem[Little and Rubin(2019)]%
        {little2019statistical}
\bibfield{author}{\bibinfo{person}{Roderick~JA Little} {and}
  \bibinfo{person}{Donald~B Rubin}.} \bibinfo{year}{2019}\natexlab{}.
\newblock \bibinfo{booktitle}{\emph{Statistical analysis with missing data}}.
  Vol.~\bibinfo{volume}{793}.
\newblock \bibinfo{publisher}{John Wiley \& Sons}.
\newblock


\bibitem[Ma et~al\mbox{.}(2018)]%
        {ma2018partial}
\bibfield{author}{\bibinfo{person}{Chao Ma}, \bibinfo{person}{Wenbo Gong},
  \bibinfo{person}{Jos{\'e}~Miguel Hern{\'a}ndez-Lobato}, \bibinfo{person}{Noam
  Koenigstein}, \bibinfo{person}{Sebastian Nowozin}, {and}
  \bibinfo{person}{Cheng Zhang}.} \bibinfo{year}{2018}\natexlab{}.
\newblock \showarticletitle{Partial VAE for hybrid recommender system}. In
  \bibinfo{booktitle}{\emph{NIPS Workshop on Bayesian Deep Learning}},
  Vol.~\bibinfo{volume}{2018}.
\newblock


\bibitem[Ma et~al\mbox{.}(2019)]%
        {ma2018eddi}
\bibfield{author}{\bibinfo{person}{Chao Ma}, \bibinfo{person}{Sebastian
  Tschiatschek}, \bibinfo{person}{Konstantina Palla},
  \bibinfo{person}{Jos{\'{e}}~Miguel Hern{\'{a}}ndez{-}Lobato},
  \bibinfo{person}{Sebastian Nowozin}, {and} \bibinfo{person}{Cheng Zhang}.}
  \bibinfo{year}{2019}\natexlab{}.
\newblock \showarticletitle{{EDDI:} Efficient Dynamic Discovery of High-Value
  Information with Partial {VAE}}. In \bibinfo{booktitle}{\emph{Proceedings of
  the 36th International Conference on Machine Learning, {ICML} 2019, 9-15 June
  2019, Long Beach, California, {USA}}} \emph{(\bibinfo{series}{Proceedings of
  Machine Learning Research}, Vol.~\bibinfo{volume}{97})}.
  \bibinfo{publisher}{{PMLR}}.
\newblock


\bibitem[Ma and Zhang(2021)]%
        {ma2021identifiable}
\bibfield{author}{\bibinfo{person}{Chao Ma} {and} \bibinfo{person}{Cheng
  Zhang}.} \bibinfo{year}{2021}\natexlab{}.
\newblock \showarticletitle{Identifiable Generative Models for Missing Not at
  Random Data Imputation}.
\newblock \bibinfo{journal}{\emph{Advances in Neural Information Processing
  Systems}}  \bibinfo{volume}{34} (\bibinfo{year}{2021}).
\newblock


\bibitem[Ma and Chen(2019)]%
        {ma2019missing}
\bibfield{author}{\bibinfo{person}{Wei Ma} {and} \bibinfo{person}{George~H.
  Chen}.} \bibinfo{year}{2019}\natexlab{}.
\newblock \showarticletitle{Missing Not at Random in Matrix Completion: The
  Effectiveness of Estimating Missingness Probabilities Under a Low Nuclear
  Norm Assumption}. In \bibinfo{booktitle}{\emph{Advances in Neural Information
  Processing Systems 32: Annual Conference on Neural Information Processing
  Systems 2019, NeurIPS 2019, December 8-14, 2019, Vancouver, BC, Canada}}.
\newblock


\bibitem[Mattei and Frellsen(2019)]%
        {mattei2019miwae}
\bibfield{author}{\bibinfo{person}{Pierre{-}Alexandre Mattei} {and}
  \bibinfo{person}{Jes Frellsen}.} \bibinfo{year}{2019}\natexlab{}.
\newblock \showarticletitle{{MIWAE:} Deep Generative Modelling and Imputation
  of Incomplete Data Sets}. In \bibinfo{booktitle}{\emph{Proceedings of the
  36th International Conference on Machine Learning, {ICML} 2019, 9-15 June
  2019, Long Beach, California, {USA}}} \emph{(\bibinfo{series}{Proceedings of
  Machine Learning Research}, Vol.~\bibinfo{volume}{97})}.
  \bibinfo{publisher}{{PMLR}}.
\newblock


\bibitem[Muzellec et~al\mbox{.}(2020)]%
        {muzellec2020missing}
\bibfield{author}{\bibinfo{person}{Boris Muzellec}, \bibinfo{person}{Julie
  Josse}, \bibinfo{person}{Claire Boyer}, {and} \bibinfo{person}{Marco
  Cuturi}.} \bibinfo{year}{2020}\natexlab{}.
\newblock \showarticletitle{Missing data imputation using optimal transport}.
  In \bibinfo{booktitle}{\emph{International Conference on Machine Learning}}.
  PMLR.
\newblock


\bibitem[Nazabal et~al\mbox{.}(2020)]%
        {nazabal2020handling}
\bibfield{author}{\bibinfo{person}{Alfredo Nazabal}, \bibinfo{person}{Pablo~M
  Olmos}, \bibinfo{person}{Zoubin Ghahramani}, {and} \bibinfo{person}{Isabel
  Valera}.} \bibinfo{year}{2020}\natexlab{}.
\newblock \showarticletitle{Handling incomplete heterogeneous data using vaes}.
\newblock \bibinfo{journal}{\emph{Pattern Recognition}}  \bibinfo{volume}{107}
  (\bibinfo{year}{2020}).
\newblock


\bibitem[Rubin(1976)]%
        {rubin1976inference}
\bibfield{author}{\bibinfo{person}{Donald~B Rubin}.}
  \bibinfo{year}{1976}\natexlab{}.
\newblock \showarticletitle{Inference and missing data}.
\newblock \bibinfo{journal}{\emph{Biometrika}} \bibinfo{volume}{63},
  \bibinfo{number}{3} (\bibinfo{year}{1976}).
\newblock


\bibitem[Saito(2020)]%
        {saito2020asymmetric}
\bibfield{author}{\bibinfo{person}{Yuta Saito}.}
  \bibinfo{year}{2020}\natexlab{}.
\newblock \showarticletitle{Asymmetric Tri-training for Debiasing
  Missing-Not-At-Random Explicit Feedback}. In
  \bibinfo{booktitle}{\emph{Proceedings of the 43rd International {ACM} {SIGIR}
  conference on research and development in Information Retrieval, {SIGIR}
  2020, Virtual Event, China, July 25-30, 2020}}. \bibinfo{publisher}{{ACM}}.
\newblock
\urldef\tempurl%
\url{https://doi.org/10.1145/3397271.3401114}
\showDOI{\tempurl}


\bibitem[Schnabel et~al\mbox{.}(2016)]%
        {schnabel2016recommendations}
\bibfield{author}{\bibinfo{person}{Tobias Schnabel}, \bibinfo{person}{Adith
  Swaminathan}, \bibinfo{person}{Ashudeep Singh}, \bibinfo{person}{Navin
  Chandak}, {and} \bibinfo{person}{Thorsten Joachims}.}
  \bibinfo{year}{2016}\natexlab{}.
\newblock \showarticletitle{Recommendations as Treatments: Debiasing Learning
  and Evaluation}. In \bibinfo{booktitle}{\emph{Proceedings of the 33nd
  International Conference on Machine Learning, {ICML} 2016, New York City, NY,
  USA, June 19-24, 2016}} \emph{(\bibinfo{series}{{JMLR} Workshop and
  Conference Proceedings}, Vol.~\bibinfo{volume}{48})}.
  \bibinfo{publisher}{JMLR.org}.
\newblock


\bibitem[Song and Ermon(2019)]%
        {song2019generative}
\bibfield{author}{\bibinfo{person}{Yang Song} {and} \bibinfo{person}{Stefano
  Ermon}.} \bibinfo{year}{2019}\natexlab{}.
\newblock \showarticletitle{Generative Modeling by Estimating Gradients of the
  Data Distribution}. In \bibinfo{booktitle}{\emph{Advances in Neural
  Information Processing Systems 32: Annual Conference on Neural Information
  Processing Systems 2019, NeurIPS 2019, December 8-14, 2019, Vancouver, BC,
  Canada}}.
\newblock


\bibitem[Song et~al\mbox{.}(2020)]%
        {song2020score}
\bibfield{author}{\bibinfo{person}{Yang Song}, \bibinfo{person}{Jascha
  Sohl-Dickstein}, \bibinfo{person}{Diederik~P Kingma},
  \bibinfo{person}{Abhishek Kumar}, \bibinfo{person}{Stefano Ermon}, {and}
  \bibinfo{person}{Ben Poole}.} \bibinfo{year}{2020}\natexlab{}.
\newblock \showarticletitle{Score-based generative modeling through stochastic
  differential equations}.
\newblock \bibinfo{journal}{\emph{arXiv preprint arXiv:2011.13456}}
  (\bibinfo{year}{2020}).
\newblock


\bibitem[Sportisse et~al\mbox{.}(2020)]%
        {sportisse2020imputation}
\bibfield{author}{\bibinfo{person}{Aude Sportisse}, \bibinfo{person}{Claire
  Boyer}, {and} \bibinfo{person}{Julie Josse}.}
  \bibinfo{year}{2020}\natexlab{}.
\newblock \showarticletitle{Imputation and low-rank estimation with missing not
  at random data}.
\newblock \bibinfo{journal}{\emph{Statistics and Computing}}
  \bibinfo{volume}{30}, \bibinfo{number}{6} (\bibinfo{year}{2020}).
\newblock


\bibitem[Steck(2013)]%
        {steck2013evaluation}
\bibfield{author}{\bibinfo{person}{Harald Steck}.}
  \bibinfo{year}{2013}\natexlab{}.
\newblock \showarticletitle{Evaluation of recommendations: rating-prediction
  and ranking}. In \bibinfo{booktitle}{\emph{Proceedings of the 7th ACM
  conference on Recommender systems}}. \bibinfo{pages}{213--220}.
\newblock


\bibitem[Stekhoven and B{\"u}hlmann(2012)]%
        {stekhoven2012missforest}
\bibfield{author}{\bibinfo{person}{Daniel~J Stekhoven} {and}
  \bibinfo{person}{Peter B{\"u}hlmann}.} \bibinfo{year}{2012}\natexlab{}.
\newblock \showarticletitle{MissForest—non-parametric missing value
  imputation for mixed-type data}.
\newblock \bibinfo{journal}{\emph{Bioinformatics}} \bibinfo{volume}{28},
  \bibinfo{number}{1} (\bibinfo{year}{2012}).
\newblock


\bibitem[Tashiro et~al\mbox{.}(2021)]%
        {tashiro2021csdi}
\bibfield{author}{\bibinfo{person}{Yusuke Tashiro}, \bibinfo{person}{Jiaming
  Song}, \bibinfo{person}{Yang Song}, {and} \bibinfo{person}{Stefano Ermon}.}
  \bibinfo{year}{2021}\natexlab{}.
\newblock \showarticletitle{CSDI: Conditional score-based diffusion models for
  probabilistic time series imputation}.
\newblock \bibinfo{journal}{\emph{Advances in Neural Information Processing
  Systems}}  \bibinfo{volume}{34} (\bibinfo{year}{2021}),
  \bibinfo{pages}{24804--24816}.
\newblock


\bibitem[Van~Buuren and Groothuis-Oudshoorn(2011)]%
        {van2011mice}
\bibfield{author}{\bibinfo{person}{Stef Van~Buuren} {and}
  \bibinfo{person}{Karin Groothuis-Oudshoorn}.}
  \bibinfo{year}{2011}\natexlab{}.
\newblock \showarticletitle{mice: Multivariate imputation by chained equations
  in R}.
\newblock \bibinfo{journal}{\emph{Journal of statistical software}}
  \bibinfo{volume}{45} (\bibinfo{year}{2011}).
\newblock


\bibitem[Wang et~al\mbox{.}(2019)]%
        {wang2019doubly}
\bibfield{author}{\bibinfo{person}{Xiaojie Wang}, \bibinfo{person}{Rui Zhang},
  \bibinfo{person}{Yu Sun}, {and} \bibinfo{person}{Jianzhong Qi}.}
  \bibinfo{year}{2019}\natexlab{}.
\newblock \showarticletitle{Doubly Robust Joint Learning for Recommendation on
  Data Missing Not at Random}. In \bibinfo{booktitle}{\emph{Proceedings of the
  36th International Conference on Machine Learning, {ICML} 2019, 9-15 June
  2019, Long Beach, California, {USA}}} \emph{(\bibinfo{series}{Proceedings of
  Machine Learning Research}, Vol.~\bibinfo{volume}{97})}.
  \bibinfo{publisher}{{PMLR}}.
\newblock


\bibitem[Wang and Blei(2019)]%
        {wang2019blessings}
\bibfield{author}{\bibinfo{person}{Yixin Wang} {and} \bibinfo{person}{David~M
  Blei}.} \bibinfo{year}{2019}\natexlab{}.
\newblock \showarticletitle{The blessings of multiple causes}.
\newblock \bibinfo{journal}{\emph{J. Amer. Statist. Assoc.}}
  \bibinfo{volume}{114}, \bibinfo{number}{528} (\bibinfo{year}{2019}).
\newblock


\bibitem[Wang et~al\mbox{.}(2020)]%
        {wang2020information}
\bibfield{author}{\bibinfo{person}{Zifeng Wang}, \bibinfo{person}{Xi Chen},
  \bibinfo{person}{Rui Wen}, \bibinfo{person}{Shao-Lun Huang},
  \bibinfo{person}{Ercan Kuruoglu}, {and} \bibinfo{person}{Yefeng Zheng}.}
  \bibinfo{year}{2020}\natexlab{}.
\newblock \showarticletitle{Information theoretic counterfactual learning from
  missing-not-at-random feedback}.
\newblock \bibinfo{journal}{\emph{Advances in Neural Information Processing
  Systems}}  \bibinfo{volume}{33} (\bibinfo{year}{2020}).
\newblock


\bibitem[Yoon et~al\mbox{.}(2018)]%
        {yoon2018gain}
\bibfield{author}{\bibinfo{person}{Jinsung Yoon}, \bibinfo{person}{James
  Jordon}, {and} \bibinfo{person}{Mihaela van~der Schaar}.}
  \bibinfo{year}{2018}\natexlab{}.
\newblock \showarticletitle{{GAIN:} Missing Data Imputation using Generative
  Adversarial Nets}. In \bibinfo{booktitle}{\emph{Proceedings of the 35th
  International Conference on Machine Learning, {ICML} 2018,
  Stockholmsm{\"{a}}ssan, Stockholm, Sweden, July 10-15, 2018}}
  \emph{(\bibinfo{series}{Proceedings of Machine Learning Research},
  Vol.~\bibinfo{volume}{80})}. \bibinfo{publisher}{{PMLR}}.
\newblock


\end{thebibliography}
\newpage
%%
%% If your work has an appendix, this is the place to put it.
\appendix

\end{document}